\documentclass{article} % For LaTeX2e
\usepackage[final]{colm2026_conference}   % camera-ready

\usepackage{microtype}
\usepackage{subcaption}
\usepackage{booktabs} % for professional tables
\usepackage{caption}
\usepackage{adjustbox}
\usepackage{url}
\usepackage{hyperref}
\usepackage[table]{xcolor}
\usepackage{soul}

\usepackage{booktabs}
\usepackage{multirow}
\usepackage{graphicx}
\usepackage{fontawesome5} % for the \faGithub icon in the author block
\usepackage{amsmath}
\usepackage{amssymb}
\usepackage{mathtools}
\usepackage{amsthm}

\usepackage[capitalize,noabbrev]{cleveref}

\theoremstyle{plain}

\theoremstyle{definition}

\theoremstyle{remark}

\usepackage{url}
\usepackage{booktabs}

\usepackage{lineno}

\definecolor{darkblue}{rgb}{0, 0, 0.5}
\hypersetup{colorlinks=true, citecolor=darkblue, linkcolor=darkblue, urlcolor=darkblue}

\definecolor{darkred}{rgb}{0.55, 0.0, 0.0}
\title{Preserving Long-Tailed Expert Information in Mixture-of-Experts Tuning}
\author{Haoze He$^{1}$ \enspace
Xingyuan Ding$^{1}$ \enspace
Xuan Jiang$^{2}$ \enspace
Xinkai Zou$^{3}$ \enspace
Alex Cheng$^{1}$ \enspace
Yibo Zhao$^{4}$ \\[2pt]
\textbf{Juncheng Billy Li}$^{1}$\thanks{Co-advisor.} \quad
\textbf{Heather Miller}$^{1}$\footnotemark[1] \\[6pt]
$^{1}$Carnegie Mellon University \quad
$^{2}$MIT \quad
$^{3}$UCSD \quad
$^{4}$Johns Hopkins University \\[4pt]
\texttt{\{haozeh, xingyuad, abcheng, junchenl, heather.miller\}@cs.cmu.edu} \\
\texttt{xuanj@mit.edu} \quad
\texttt{x9zou@ucsd.edu} \quad
\texttt{yzhao231@jh.edu} \\[4pt]
{\small\faGithub}~\url{https://github.com/HectorHHZ/ExpertCondenser}
}

\begin{document}

\ifcolmsubmission
\linenumbers
\fi

\maketitle

\begin{abstract}
Despite MoE models leading many benchmarks, supervised fine-tuning (SFT) for the MoE architectures remains difficult because its router layers are fragile.
Methods such as DenseMixer and ESFT mitigate router collapse with dense mixing or auxiliary load-balancing losses, but these introduce noisy gradients that often degrade performance.
In preliminary experiments, we systematically pruned experts and observed that while certain “super experts” are activated far more frequently, discarding less used experts still leads to notable performance degradation. This suggests that even rarely activated experts encode non-trivial knowledge useful for downstream tasks. 
Motivated by this, we propose an auxiliary-loss-free MoE SFT framework that combines bias-driven sparsification with always-active gated condenser experts. Rather than enforcing balanced activation across all experts, our method encourages task-relevant experts to remain active while pushing long-tailed experts toward inactivity. The condenser experts provide a persistent, learnable pathway that alleviates gradient starvation and facilitates consolidation of information that would otherwise remain fragmented across sparsely activated experts. Analysis further suggest that this design better preserves long-tailed expert information under sparse routing.
Experiments on large-scale MoE models demonstrate that our approach outperforms state-of-the-art SFT baselines such as DenseMixer and ESFT, achieving average gain of 2.5\%+ on both mathematical reasoning and commonsenseQA benchmarks.
% Motivated by this,
% we propose a new auxiliary-loss-free MoE SFT framework that combines router biases with shared condenser experts.
% Instead of enforcing balanced activation across all experts, our method leverages bias updates to encourage imbalanced and sparse routing, allowing rarely used experts to become inactive while designating existing experts as shared condenser experts that aggregate knowledge from the inactive set without increasing the per-token compute budget.
\end{abstract}

\section {Introduction}

Mixture-of-Experts (MoE) models scale language models efficiently by activating only a small subset of experts per token, enabling massive capacity without increasing per-token compute. Yet the same sparse routing that drives their success also makes them fragile: MoE relies on a non-differentiable Top-K router, which blocks straightforward gradient flow and makes post-training, such as supervised fine-tuning (SFT), far more difficult than for dense LLMs.

Over the years, researchers have sought to stabilize MoE training through progressively refined routing strategies. GShard~\citep{lepikhin2020gshard} introduced top-2 gating with heavy auxiliary balancing losses, while Switch Transformers~\citep{fedus2022switch} simplified this to a single expert per token. More recent work, such as DeepSeek-MoE~\citep{wang2024auxiliary} and DeepSeek-V3~\citep{liu2024deepseek3}, explored bias-based routers and minimized auxiliary losses to reduce gradient noise and improve efficiency. However, these advances primarily address pre-training. In the post-training setting, SFT remains underexplored: ESFT~\citep{wang2024let} routes all gradients to the most-activated expert, while DenseMixer~\citep{yao2025densemixer} improves slightly by applying a Straight-Through Estimator (STE) \citep{bengio2013estimating} to approximate updates for inactive experts, yet STE introduces biased gradients.

In parallel, recent studies have identified the existence of “super experts”\cite{su2025unveiling} or “super weights”\cite{super-weight}, whose activations dominate the routing and whose removal leads to sharp performance degradation. These findings suggest that a small subset of experts carries disproportionate importance. However, our observations reveal a complementary phenomenon: even the rarely activated experts encode indispensable information, and pruning them also causes substantial performance decline. This highlights the need for fine-tuning strategies that not only preserve the capacity of frequently activated super experts, but also consolidate information from the long tail of rarely activated experts.

Motivated by these observations, we 
introduce \textbf{condenser experts}: experts that are always selected yet remain gated per token.
% This design provides a persistent yet input-dependent computation pathway, allowing the model to better preserve and consolidate information that would otherwise remain fragmented across rarely activated experts.
% adapt \bl{todo}the bias-updating principle of DeepSeek’s Loss-Free Balancing to the post-training setting. Instead of aiming for balanced activation across all experts, we propose an auxiliary-free fine-tuning framework that enforces sparse routing through globally negative biases. 
% This drives rarely used experts toward inactivity, while two designated condenser experts remain active and receive persistent gradient updates alongside routed experts. 
This creates a learnable shared pathway that alleviates gradient starvation, narrows the train–inference routing gap, and facilitates preservation of information that would otherwise remain distributed across long-tailed experts.
Experimental results show that our method consistently outperforms SoTA MoE SFT methods by 2.5\%+ when post-training popular MoE LLMs on commonsense reasoning (PIQA, ARC, SIQA) and math reasoning benchmarks (MATH-500, AIME-25, GPQA, GSM8K, etc). Our implementation is open-source.

%Anonymously~\footnote{\url{https://anonymous.4open.science/r/Finetuning-MOE-F652}}.

% \bl{meta todos: 1. ablation of number of condensor expert seems necessary. 1,2,4,8. Condenser experts act like a low-rank bottleneck over expert space. the number of the condensor experts mimicks rank; 2. always-on vs. gated, why always-on is better?}

\subsection{Contributions}

\begin{itemize}
    \item We extend scaling-law analysis to MoE compression by relating performance to the number of expert parameters retained. Through this analysis, we find that performance degrades substantially when rarely activated experts are removed, showing that these long-tailed experts encode indispensable knowledge beyond the most frequently activated ``super experts.''
    \item We propose \textbf{ExpertCondenser}, a MoE post-training method combines auxiliary-free bias-based routing with always-active gated condenser experts to alleviate gradient starvation, narrow the train–inference routing gap, and facilitate consolidation of information from long-tailed experts.
    
    % \bl{makes the claims simple: 1. prevents gradient starvation; 2. preserves train-inference consistency better than dense mixing.}
    
\end{itemize}

% \bl{need to discuss\cite{yao2025densemixer} is eventually using dense model a good idea?
% what aobut \cite{wang2025two} two experts, this is also bold claim}
% \newpage
% \clearpage
\section{Does Saving ``Super Experts'' Mean Saving Model Performance?}
\label{sec:scaling_laws}

\begin{figure*}[t] 
    \vspace{-0.5in}
    \centering
    \includegraphics[width=1.\textwidth]{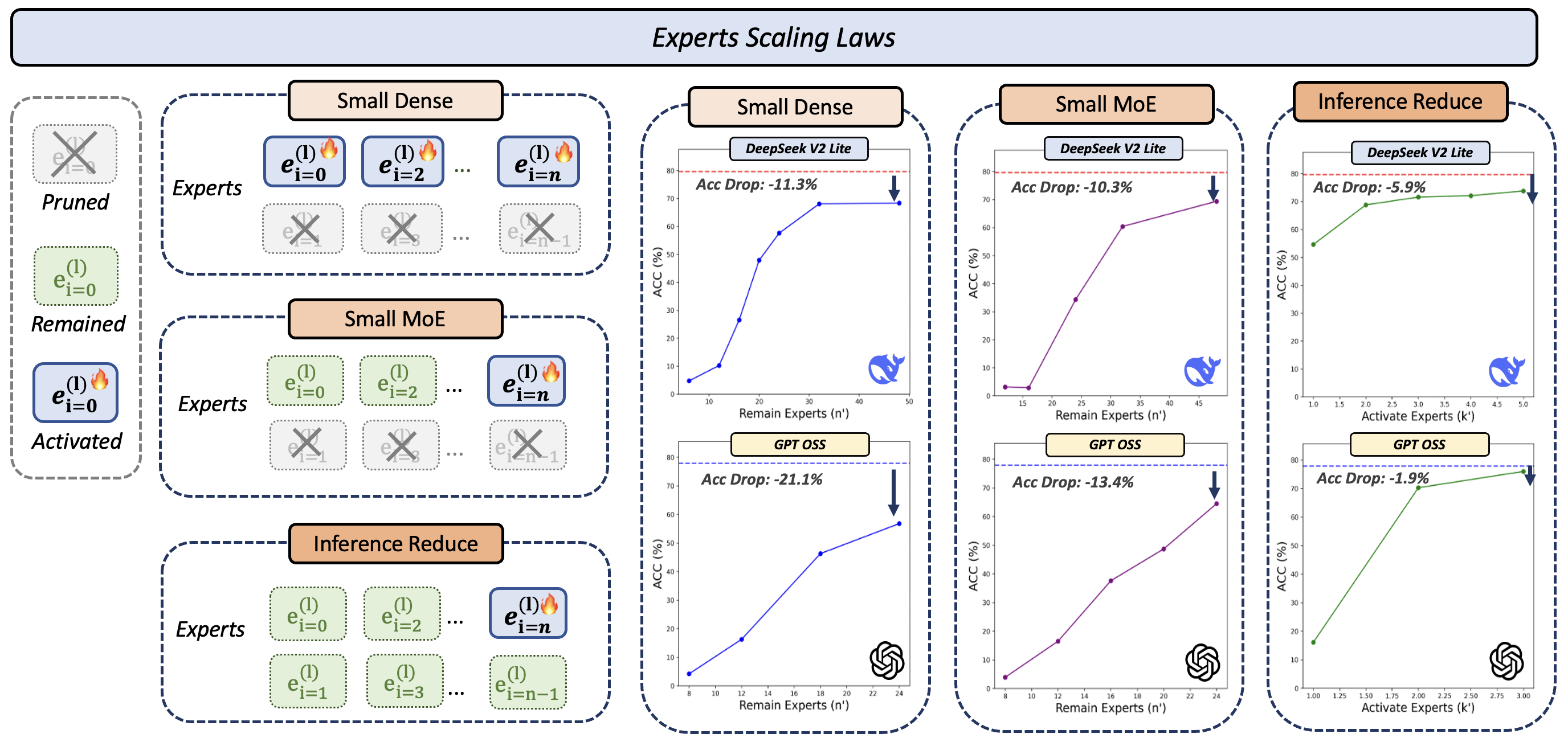} 
    \vspace{-0.2in}
    \caption{Illustration of the three expert scaling strategies. 
    (\textit{Small-Dense}) Experts are pruned, and all remaining experts are always activated. 
    (\textit{Small MoE}) Experts are pruned and only a subset of them are activated per token. 
    (\textit{Inference Reduce}) Fewer experts are selected per token while the expert pool is unchanged. 
    A substantial performance gap remains between the base model and pruned variants under Small Dense and small MoE categories.}
    \vspace{-0.2in}
    \label{fig:scaling_laws}
\end{figure*}

Previous work~\citep{lu2024not, su2025unveiling} shows that pruning frequently activated ``super experts'' leads to significant performance degradation, highlighting their importance. However, it remains unclear whether retaining only these experts is sufficient to preserve model quality.

To answer this, we study how performance scales with the total number of retained expert parameters. We consider three compression strategies (Fig.~\ref{fig:scaling_laws}) that vary the expert pool size $n$ and activation budget $k$. For token $t$, the selected experts are
\[
S_t = \mathrm{TopK}(\{s_{j,t}\}_{j=1}^n, k),
\]
where $s_{j,t}$ denotes the gating score.

We examine:
(i) \textbf{Dense conversion via pruning:} $n \to n'$, $k = n'$, activating all remaining experts (dense model); (ii) \textbf{Smaller MoE via pruning:} $n \to n'$, $k < n'$, preserving sparsity within a reduced expert pool; (iii) \textbf{Reduced activation budget:} $n$ fixed, $k \to k' < k$, increasing sparsity without changing model size.
We use ES-Act and ES-Mag~\citep{wang2024let} to select experts prior to pruning, with ES-Act as default (details in Appendix~\ref{appendix:experts_selection}).

To conduct a scaling-law analysis of experts, we design experiments using GPT-OSS~\citep{openai2025gptoss120bgptoss20bmodel} and DeepSeek-Coder-V2-Lite~\citep{liu2024deepseek}. In figure~\ref{fig:scaling_laws}, we summarize the results across the three strategies (more details in Appendix~\ref{appendix:scale_law_tables}, Tab.~\ref{tab:scale_laws_gpt} and \ref{tab:scale_laws_deepseek}).
Even though expert activation is highly skewed and a few “super experts” dominate routing (shown in Appendix~\ref{appendix:activation_analysis}, Fig.~\ref{fig:deepseek_base_expert_activate}), a substantial performance gap still remains between the base model and pruned variants. Retaining the top 75\% of experts still results in more than a 10\% drop. \textbf{These results show that preserving ``Super Experts'' is not sufficient to preserve model performance.}
Crucially, our experiments reveal that rarely activated experts nonetheless encode indispensable domain knowledge, which is distributed in a fragmented manner across long-tailed experts.

\section{Proposed Method}

% Our findings in Section~\ref{sec:scaling_laws} suggest that effective post-training should preserve domain knowledge from infrequently activated experts.
% % This requires a mechanism that aggregates knowledge from long-tail experts into a smaller set of always active components, while maintaining the sparse structure of MoE models.
% This motivates a mechanism that preserves information from long-tail experts while concentrating optimization into a small set of persistent components.

% While traditional shared experts provide persistent computation, they are ungated and therefore mix signals uniformly across inputs, limiting their expressive ability. More generally, persistent computation paths are most expressive when they remain learnably mixed rather than fixed (as we further discussed in Appendix~\ref{appendix: LPM}).

% Motivated by this, we introduce \textbf{condenser experts}: experts that are always selected yet remain gated per token.
% This design provides a persistent yet input-dependent computation pathway, allowing the model to better preserve and consolidate information that would otherwise remain fragmented across rarely activated experts. We refer to the overall training framework as \textbf{ExpertCondenser}.
Our findings in Section~\ref{sec:scaling_laws} suggest that effective post-training should preserve knowledge from infrequently activated experts. This motivates a mechanism that consolidates such long-tailed information into a small set of persistent components.

% While traditional shared experts provide persistent computation, they are ungated and mix signals uniformly across inputs, limiting expressiveness. In contrast, persistent computation paths are more effective when they remain input-dependent.

While traditional shared experts provide persistent computation, they are ungated and mix signals uniformly across inputs, limiting expressiveness. More generally, persistent computation paths are more expressive when they remain learnably mixed rather than fixed. A detailed discussion between condenser experts and shared experts is in Appendix~\ref{appendix: LPM}.

We therefore introduce \textbf{condenser experts}: experts that are always selected yet remain gated per token. This design enables a persistent but input-adaptive pathway, allowing the model to better preserve and consolidate information that would otherwise remain fragmented across rarely activated experts. We refer to the overall training framework as \textbf{ExpertCondenser}.

\subsection{Condenser Experts}
\label{subsection:method_condenser_experts}

\begin{figure*}[t!] 
    \vspace{-0.4in}
    \centering
    \includegraphics[width=0.8\linewidth]{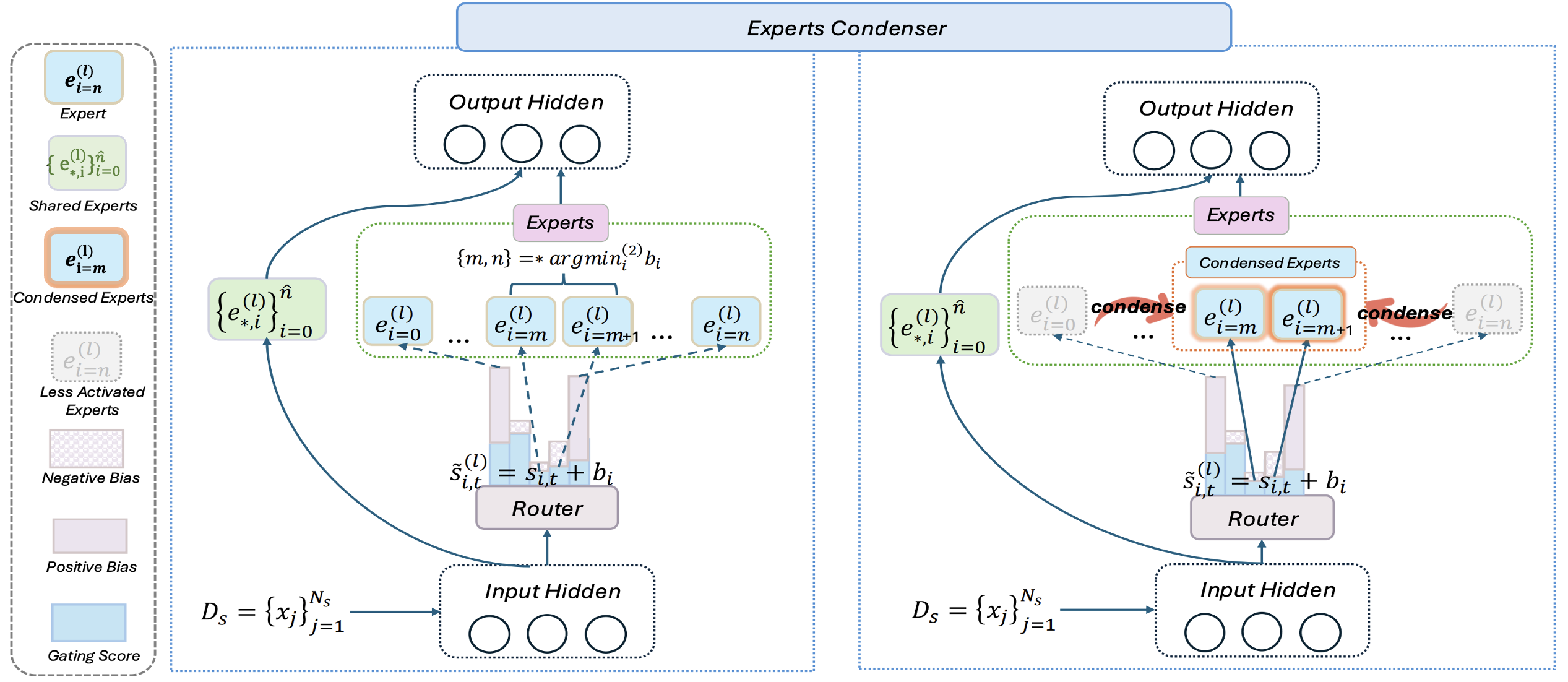}
    \caption{Representation of our Experts Condenser framework. 
    \textit{(Left)} An auxiliary-free router adds trainable biases $b_i$ to logits $s_{i,t}$. Less relevant experts accumulate negative biases, and a small set of experts are designated as condenser experts. \textit{(Right)} These condenser experts are always selected during training, ensuring they receive gradients and act as repositories that condense knowledge from inactive experts—preserving information while enforcing sparsity.}
    \vspace{-0.2in}
    \label{fig:expertcondenser}
\end{figure*}

In a standard MoE layer, for token $\mathbf{x}_t$, the output is typically written as
\begin{equation}
\mathbf{h}_t
=
\underbrace{\sum_{i=1}^{n}
g_i(\mathbf{x}_t)\, E_i(\mathbf{x}_t)}_{\text{sparse experts}}
+
\underbrace{\sum_{k=1}^{\hat{n}} E_k(\mathbf{x}_t)}_{\text{static shared experts}},
\label{eq:moe_standard}
\end{equation}
where $E(\cdot)$ are expert functions, $g(\cdot)$ are computed gate weight from the router, and $x_t$ denote token $t$.
The static shared experts are always active with fixed weight, shown as green boxes in Fig.~\ref{fig:expertcondenser}. 

 We introduce condenser experts, that are \textbf{always activated yet dynamically weighted} shown in Fig.~\ref{fig:expertcondenser}.  Let $J \subset \{1, \dots, n\}$ denote the fixed set of condenser experts with $|J| = r$ and $\tilde{s}_i(x_t)$ denote the routing score for expert $i$. We discuss the condenser experts selection in \S~\ref{subsection:Auxiliary-Free-Sparsity-Enforcement}. For each token $x_t$, we define the active expert set as
\begin{equation}
S(x_t) = J \cup \operatorname{TopK}_{i \notin J} \left( \tilde{s}_i(x_t),\, k - r \right).
\end{equation}

We reformulate the MoE output as
\begin{equation}
\label{formula:reformulate_moe}
h'_t =  
\underbrace{\sum_{j \in J} g_j(x_t)\, E_j(x_t)}_{\text{gated condenser experts}}
+
\underbrace{\sum_{i \in S(x_t)\setminus J} g_i(x_t)\, E_i(x_t)}_{\text{selected sparse experts}}
+
\underbrace{\sum_{k=1}^{\hat{n}} E_k(\mathbf{x}_t)}_{\text{static shared experts}}
\end{equation}

\paragraph{Condenser Experts Avoid Gradient Bottleneck Introduced by Router.}
\label{assumption:avoid_gradient_bottleneck}
Let
\[
Z_i(x)
=
\frac{\partial \ell}{\partial h(x)}
\cdot
g_i(x)\,
\frac{\partial E_i(x)}{\partial \theta_i}.
\]
For a routed expert $i \notin J$, the gradient can be written as
$
\nabla_{\theta_i} \mathcal{L}
=
\mathbb{E}\big[\mathbf{1}\{i \in S(x)\} \, Z_i(x)\big].
$
Let $p_i = \mathbb{P}(i \in S(x))$. By conditioning on the routing event,
$
\nabla_{\theta_i} \mathcal{L}
=
p_i \, \mathbb{E}\big[ Z_i(x) \mid i \in S(x) \big],
$
showing that routing induces a multiplicative attenuation by $p_i < 1$ on expert-specific gradient flow.
% This attenuation also manifests at the level of training dynamics. Let $B_t^{(i)} = \mathbf{1}\{i \in S(x_t)\}$ denote whether expert $i$ is selected at step $t$, and define the total number of updates over $T$ steps as
% \[
% N_i(T) = \sum_{t=1}^T B_t^{(i)}.
% \]
% Assuming independent routing with probability $p_i$, we have $N_i(T) \sim \mathrm{Binomial}(T, p_i)$. By a Chernoff bound, for any $0 < \delta < 1$,
% \begin{equation}
% \Pr\!\left(N_i(T) \le (1-\delta) T p_i \right)
% \le
% \exp\!\left(-\frac{\delta^2}{2} T p_i \right).
% \end{equation}
This attenuation also manifests at the level of training dynamics. Let 
$B_t^{(i)} = \mathbf{1}\{i \in S(x_t)\}$ denote whether expert $i$ is selected at step $t$, and define the total number of updates over $T$ steps as
$N_i(T) = \sum_{t=1}^T B_t^{(i)}$.
Let $p_i = \mathbb{P}(i \in S(x))$ denote the marginal activation probability of expert $i$. Then
$\mathbb{E}[N_i(T)] = T p_i$,
so the effective number of updates scales linearly with $p_i$. In practice, routing distributions are highly non-uniform (Figure \ref{fig:deepseek_base_expert_activate}), and $p_i$ can vary significantly across experts, leading to substantial imbalance in training exposure.
Although routing decisions are input-dependent and not strictly independent, $N_i(T)$ behaves similarly to a sum of Bernoulli trials with mean $T p_i$. By a Chernoff bound, for any $0 < \delta < 1$,
\begin{equation}
\Pr\!\left(N_i(T) \le (1-\delta) T p_i \right)
\le
\exp\!\left(-\frac{\delta^2}{2} T p_i \right).
\end{equation}
Thus, with high probability, routed experts receive only $\Theta(T p_i)$ effective updates over training.
Under mild conditions, $N_i(T)$ concentrates around its mean, so long-tailed experts consistently receive only $\Theta(T p_i)$ updates over training.

In contrast, for a condenser expert $j \in J$, we have
$\nabla_{\theta_j} \mathcal{L} = \mathbb{E}\big[ Z_j(x) \big]$,
so condenser experts avoid both the per-step gradient attenuation and the reduced effective sample size induced by routing sparsity. This highlights that gradient starvation is fundamentally a multiplicative bottleneck in both gradient magnitude and update frequency, which is eliminated by always-active condenser experts. In ~\S~\ref{subsection:routing_and_output_analysis} we show that ExpertCondenser achieve more substantial differences of MoE outputs than SFT, and in ~\S~\ref{subsection:gradient_norm_analysis}, we show the gradient norm condenser experts is consistently larger than routed experts during training, suggesting condenser expert avoid gradient bottleneck.

\paragraph{Condenser Experts Consolidate Domain Information.}
\label{assumption:consolidate_info}
Condenser experts receive gradients that reflect domain-specific patterns by co-activated with routed experts. Assume the training distribution is a mixture $\mathcal{D} = \sum_{d=1}^m \pi_d \mathcal{D}_d$. Then for a condenser expert $j \in J$,
\begin{equation}
\nabla_{\theta_j} \mathcal{L}
=
\sum_{d=1}^m \pi_d
\mathbb{E}_{(x,y)\sim \mathcal{D}_d}
\left[
\frac{\partial \ell}{\partial h(x)}
\cdot
g_j(x)\,
\frac{\partial E_j(x)}{\partial \theta_j}
\right],
\end{equation}
This shows that condenser experts accumulate gradient contributions across all domains, and integrates signals that are otherwise distributed across sparsely activated experts (more details in Appendix~\ref{appendix:condenser_theory}). Given the decomposition in Eq.~\ref{formula:reformulate_moe}, the gated condenser expert term can be represented as a weighted low-dimensional shared representation that captures common structure across routed experts. Our analysis in ~\S~\ref{subsection:condenser_experts_analysis} empirically suggesting condenser experts consolidate in-domain information.

% The size of $J$ controls a trade-off: larger $|J|$ increases shared capacity but reduces compression, while smaller $|J|$ enforces stronger consolidation of distributed knowledge into a few experts.

\subsection{Auxiliary Free Sparsity Enforcement}
\label{subsection:Auxiliary-Free-Sparsity-Enforcement}
Auxiliary losses inject competing gradient signals that can hinder convergence on specialized downstream tasks.~\footnote{More details about auxiliary losses can be found in Appendix~\ref{appendix:aux_loss} and Appendix~\ref{appendix:Auxiliary_Loss_Free}.} 
We eliminate this issue by adopting an \textbf{auxiliary-loss-free} routing strategy for fine-tuning, where the routing logits are directly modified with trainable bias parameters ${b_i}$, one per expert.
We define the router's biased logits as
\begin{equation}
\tilde{s}_i(x_t) = s_i(x_t) + b_i.
\end{equation}
where $s_i$ is the original raw logit for expert $i$. Following the DeepSeek-2/3 settings~\citep{wang2024auxiliary}, final gating weights $\rho_i$ are still computed from original unbiased logits $s_i$. ~\footnote{Find additional discussion on bias-based routing design and expert selection in Appendix~\ref{appendix:bias_selection}.}

Our objective in fine-tuning is to induce task-adaptive sparsification while preserving useful information carried by rarely activated experts.
Biases for experts that are less useful for the target task are progressively decreased, separating the expert pool into a task-relevant active set and a long tail of weakly selected experts.
The bias mechanism narrows the train–inference routing gap. In ~\S~\ref{subsection:routing_and_output_analysis}, we empirically show that the bias mechanism avoid heavily perturbing fragile routers and can further encourage stable training.

\paragraph{Tradeoff in the number of condenser experts.}
Let $r$ denote the number of condenser experts. Increasing $r$ improves aggregation capacity but reduces gradient concentration per expert, while smaller $r$ increases concentration but may create a capacity bottleneck. 
% This tradeoff can be captured by a simple surrogate objective,
% $\Phi(r) = \alpha \log r - \beta r$,
% where the first term reflects representational capacity and the second term reflects dilution of gradient aggregation. 
This predicts an interior optimum for small $r$. In practice, we find that $r=2$ provides a balance between capacity and consolidation. 
Condenser experts are selected as the $r$ experts with the lowest bias values $b_i$, which correspond to consistently activated experts under the auxiliary-free routing dynamics.
We further validate these design choices through ablation studies: in Appendix~\ref{appendix:r_ablation}, we analyze the effect of varying $r$, and in Appendix~\ref{appendix:condenser_experts_selection}, we study different strategies for selecting condenser experts.

\section{Background and Related Works}

Post-training for Mixture-of-Experts (MoE) large language models remains relatively underexplored. 
Recent efforts have primarily focused on how to adapt experts so that they better align with downstream domains. 
Two representative approaches are Expert Supervised Fine-Tuning (ESFT)~\citep{wang2024let} and DenseMixer~\citep{yao2025densemixer}, which propose different strategies for handling gradient propagation through the non-differentiable Top-$k$ routing mechanism.  

\textbf{ESFT}~\footnote{More details about ESFT and DenseMixer are in Appendix~\ref{appendix:esft_densemixer}. }  
ESFT strengthens the role of the most frequently activated experts by routing gradients only through the Top-$k$ set $S_t$. 
Formally, if $w_{i,t}$ denotes the routing weight of expert $i$ for token $t$, then the gradient with respect to router parameters $\theta$ is approximated as
$
\nabla_\theta \mathcal{L} 
\;\approx\;
\sum_{i \in S_t} 
\Bigl(\tfrac{\partial \mathcal{L}}{\partial y_t} \cdot v_i\Bigr)
\,\frac{\partial w_{i,t}}{\partial \theta}.
$
Here $S_t = \mathrm{TopK}(\{w_{j,t}\}_{j=1}^n,k)$ is the set of selected experts, and $v_i$ is the output of expert $i$. 
Notably, ESFT adopts the same router training objective as SFT~\citep{esft_code, trl_sft_trainer}.  \\ 
\textbf{DenseMixer.}~\footnotemark[3] 
DenseMixer instead addresses the non-differentiability of Top-$k$ routing by adopting a straight-through estimator (STE). 
In this view, the backward pass ignores the hard selection and treats $\mathrm{TopK}$ as the identity map:
$
\frac{\partial\,\mathrm{TopK}(w_{1,t},\dots,w_{n,t})_i}{\partial w_{j,t}}
\;\approx\;\delta_{i,j}.
$
As a result, gradients flow to \emph{all} experts’ routing weights, not just those in $S_t$:
$
\nabla_\theta \mathcal{L}
\;\approx\;
\sum_{i=1}^n
\Bigl(\tfrac{\partial \mathcal{L}}{\partial y_t} \cdot v_i\Bigr)
\,\frac{\partial w_{i,t}}{\partial \theta}.
$
However, both ESFT~\citep{wang2024let} and DenseMixer~\citep{yao2025densemixer} still suffer from the additional noise introduced by auxiliary balancing losses. Moreover, they overlook the fact that less frequently activated experts also encode indispensable knowledge and contribute significantly to overall model performance. ESFT explicitly reinforces the ``super experts'' that dominate activation, the less activated experts are frozen and treated as trivial experts.

% Recent studies have explored various aspects of LLM distillation. For instance, \cite{liang2020mixkd} introduced MixKD, a data-agnostic distillation framework that enhances generalization by leveraging data augmentation techniques. Additionally, \cite{sun2025tinyR1} presented the Branch-Merge distillation approach, which combines specialized student models to improve performance across multiple benchmarks. These works inform our understanding of distillation techniques and motivate our focus on efficient methods tailored for MoE architectures.

\begin{table*}[t]
\centering
\vspace{-0.25in}
\caption{Evaluation of post-trained models (zero-shot results) on downstream math reasoning datasets, including SingleEq, MultiArith, AddSub, GSM8K, SVAMP, AQuA, and MAWPS. ExpertCondenser achieves relatively stronger performance on simpler arithmetic datasets such as SingleEq and AddSub, which primarily require short-horizon reasoning. }
\label{tab:results_math7k_14k}
\resizebox{\textwidth}{!}{%
\setlength{\tabcolsep}{1.5mm}%
\begin{tabular}{p{1.4cm} p{2.4cm} c c c c c c c c c c}
\toprule
\multirow{2}{*}{\textbf{Dataset}} 
& \multirow{2}{*}{\textbf{Model}} 
& \multirow{2}{*}{\textbf{Model Size}} 
& \multirow{2}{*}{\textbf{Post-train Type}}
& \multirow{2}{*}{\textbf{GSM8K}}
& \multirow{2}{*}{\textbf{SingleEq}}
& \multirow{2}{*}{\textbf{SVAMP}} 
& \multirow{2}{*}{\textbf{MultiArith}} 
& \multirow{2}{*}{\textbf{AddSub}} 
& \multirow{2}{*}{\textbf{AQuA}} 
& \multirow{2}{*}{\textbf{MAWPS}} 
& \multirow{2}{*}{\textbf{AVG}} \\
& & & & & & & & & & & \\
\midrule

%===== Block 1: math7k =====
\multirow{16}{*}{\textbf{math7k}} 

& \multirow{4}{*}{\textbf{Qwen1.5-MoE}} 
& \multirow{4}{*}{\textbf{14B}}
& SFT & 45.8 & 70.2 & 53.6 & 76.2 & 53.3 & 27.6 & 67.8 & 56.4 \\
& & & ESFT & 46.9 & 69.3 & 54.1 & 75.7 & 52.2 & 27.6 & 68.1 & 56.2 \\
& & & DenseMixer & 54.6 & 72.1 & 52.4 & 81.8 & 58.7 & 27.8 & 72.5 & 60.0 \\
& & & ExpertCondenser (Ours) & 57.2 & 73.6 & 55.7 & 86.0 & 61.8 & 33.1 & 75.6 & \textbf{63.3} \\
\cmidrule(l){2-12}

& \multirow{4}{*}{\textbf{DeepSeek-V2-Lite}}
& \multirow{4}{*}{\textbf{16B}}
& SFT & 58.4 & 80.6 & 66.2 & 90.2 & 61.0 & 24.8 & 75.8 & 65.3 \\
& & & ESFT & 58.6 & 80.9 & 65.8 & 90.7 & 62.3 & 27.6 & 76.1 & 66.0 \\
& & & DenseMixer & 57.6 & 88.6 & 67.8 & 92.3 & 72.8 & 33.1 & 80.6 & 70.4 \\
& & & ExpertCondenser (Ours) & 59.4 & 92.5 & 69.1 & 91.5 & 79.5 & 36.1 & 83.6 & \textbf{73.1} \\

\cmidrule(l){2-12}

& \multirow{4}{*}{\textbf{OLMoE}} 
& \multirow{4}{*}{\textbf{7B}}
& SFT            & 63.6 & 74.8 & 67.7 & 92.3 & 57.8 & 27.6 & 72.4 & 65.2 \\
& & & ESFT       & 62.2 & 75.2 & 68.0 & 93.3 & 58.2 & 28.7 & 73.1 & 65.5 \\
& & & DenseMixer & 64.6 & 78.2 & 68.4 & 92.3 & 59.2 & 36.3 & 74.2 & 67.6 \\
& & & ExpertCondenser (Ours) & 68.4 & 79.8 & 71.2 & 93.8 & 63.4 & 36.3 & 78.8 & \textbf{70.2} \\
\cmidrule(l){2-12}

& \multirow{4}{*}{\textbf{GPT-OSS}}
& \multirow{4}{*}{\textbf{20B}}
& SFT & 78.2 & 92.9 & 81.2 & 98.2 & 82.5 & 36.8 & 91.2 & 80.1 \\
& & & ESFT & 76.6 & 92.9 & 80.2 & 98.2 & 82.0 & 35.4 & 90.3 & 79.4 \\
& & & DenseMixer & 80.1 & 92.3 & 83.2 & 98.7 & 82.5 & 37.4 & 90.8 & 80.7 \\
& & & ExpertCondenser (Ours) & 81.7 & 93.2 & 82.5 & 98.5 & 85.6 & 38.6 & 91.6 & \textbf{81.7} \\

%===== Block 2: math14k =====
\midrule

\multirow{12}{*}{\textbf{math14k}}

& \multirow{4}{*}{\textbf{Qwen1.5-MoE}} 
& \multirow{4}{*}{\textbf{14B}}
& SFT & 51.8 & 74.8 & 55.4 & 87.2 & 66.7 & 27.6 & 74.8 & 62.6 \\
& & & ESFT & 52.5 & 76.0 & 54.1 & 86.2 & 62.3 & 29.5 & 71.4 & 61.7 \\
& & & DenseMixer & 55.8 & 79.6 & 57.2 & 88.7 & 70.8 & 31.6 & 76.6 & 65.8 \\
& & & ExpertCondenser (Ours) & 58.8 & 81.2 & 59.2 & 91.6 & 72.8 & 33.2 & 78.4 & \textbf{67.9} \\

\cmidrule(l){2-12}

& \multirow{4}{*}{\textbf{DeepSeek-V2-Lite}} 
& \multirow{4}{*}{\textbf{16B}}
& SFT & 57.6 & 76.4 & 67.6 & 90.1 & 59.7 & 30.6 & 74.3 & 65.2 \\
& & & ESFT & 58.2 & 75.8 & 65.2 & 89.0 & 56.5 & 29.5 & 73.5 & 64.0 \\
& & & DenseMixer & 61.6 & 79.4 & 69.6 & 91.6 & 57.8 & 28.6 & 78.7 & 66.8 \\
& & & ExpertCondenser (Ours) & 63.6 & 81.2 & 71.8 & 93.8 & 60.8 & 33.2 & 81.4 & \textbf{69.4} \\
\cmidrule(l){2-12}

& \multirow{4}{*}{\textbf{OLMoE}} 
& \multirow{4}{*}{\textbf{7B}}
& SFT & 65.3 & 78.4 & 69.8 & 82.8 & 68.5 & 31.4 & 75.8 & 67.4 \\
& & & ESFT & 64.4 & 77.0 & 68.9 & 81.8 & 64.1 & 30.7 & 74.8 & 65.9 \\
& & & DenseMixer & 66.2 & 78.4 & 68.6 & 83.6 & 66.8 & 32.8 & 74.6 & 67.3 \\
& & & ExpertCondenser (Ours) & 67.8 & 81.6 & 72.4 & 86.8 & 68.8 & 32.8 & 79.6 & \textbf{70.0} \\

\bottomrule
\end{tabular}%
}
\vspace{-0.05in}
\end{table*}

\begin{table*}[t!]
\vspace{-0.1in}
\centering
\caption{Evaluation of post-trained models (\textit{Pass@4}) on downstream math reasoning benchmarks after fine-tuning with \textit{Stanford-S1}, including GPQA Diamond, AIME 24/25, and MATH-500. Note that Qwen1.5 and DeepSeek-Coder-V2-Lite are earlier-generation MoE models with limited mathematical reasoning capacity, leading to near-zero performance on AIME 24/25 across all methods.  }
\label{tab:dataset_results_stanfords1}

\vspace{-0.1in}
\resizebox{\textwidth}{!}{%
\setlength{\tabcolsep}{3mm}%
\begin{tabular}{p{3cm} c c c c c c c}
\toprule
\multirow{2}{*}{\textbf{Model}} 
& \multirow{2}{*}{\textbf{Model Size}} 
& \multirow{2}{*}{\textbf{Distill Type}}
& \multirow{2}{*}{\textbf{GPQA Diamond}}
& \multirow{2}{*}{\textbf{AIME 2024}}
& \multirow{2}{*}{\textbf{AIME 2025}} 
& \multirow{2}{*}{\textbf{MATH-500}} 
& \multirow{2}{*}{\textbf{AVG}} \\
& & & & & & & \\
\midrule

%===== Qwen1.5 =====
\multirow{4}{*}{\textbf{Qwen1.5-MoE}} 
& \multirow{4}{*}{\textbf{14B}} 
% & Base Model & 25.9 & 0.0 & 0.0 & 8.4 & 8.6 \\
& SFT & 27.8 & 0.8(1/120) & 0.0(0/120) & 20.1 & 12.2 \\
& & ESFT & 26.4 & 0.8(1/120) & 0.8(1/120) & 18.1 & 11.5 \\
& & DenseMixer & 31.6 & 1.7(2/120) & 1.7(2/120) & 26.7 & 15.4 \\
& & ExpertCondenser (Ours) & 34.6 & 4.1(5/120) & 2.5(3/120) & 28.6 & \textbf{17.4} \\

\midrule

%===== DeepSeek =====
\multirow{4}{*}{\textbf{DeepSeek-Coder-V2-Lite}} 
& \multirow{4}{*}{\textbf{16B}}
% & Base Model & 31.9 & 0.8(1/120) & 1.7(2/120) & 62.0 & 24.1 \\
& SFT & 34.2 & 2.5(3/120) & 2.5(3/120) & 64.6 & 26.0 \\
& & ESFT & 32.2 & 2.5(3/120) & 2.5(3/120) & 63.0 & 25.0 \\
& & DenseMixer & 34.8 & 2.5(3/120) & 2.5(3/120) & 64.8 & 26.1 \\
& & ExpertCondenser (Ours) & 38.7 & 3.3(4/120) & 2.5(3/120) & 66.8 & \textbf{27.8} \\
\midrule
%===== Qwen3 =====
\multirow{4}{*}{\textbf{Qwen3}} 
& \multirow{4}{*}{\textbf{30B}} 
% & Base Model & 38.9 & 20.8(25/120) & 7.5(9/120) & 72.6 & 35.0 \\
& SFT & 58.6 & 63.3(76/120) & 48.3(58/120) & 94.8 & 66.3 \\
& & ESFT & 52.7 & 61.7(74/120) & 44.2(53/120) & 92.0 & 62.7 \\
& & DenseMixer & 61.0 & 65.8(79/120) & 46.7(56/120) & 95.8 & 67.3 \\
& & ExpertCondenser (Ours) & 65.8 & 68.3(82/120) & 50.0(60/120) & 96.8 & \textbf{70.2} \\

\bottomrule
\end{tabular}%
}
\vspace{-0.1in}
\end{table*}

\section{Experimental Results}

In this section, we investigate how ExpertCondenser compared to the state-of-the-art methods~\citep{wang2024auxiliary,yao2025densemixer} in terms of model performance and system efficiency.

To answer this, in ~\S~\ref{sec:experiment_math_reasoning} and ~\S~\ref{sec:experiment_commonsense_reasoning}, we evaluate ExpertCondenser on mathematical and commonsense reasoning benchmarks in term of model performance. In ~\S~\ref{sec:system_efficiency}, we analyze the system efficiency of ExpertCondenser in terms of training system efficiency. We evaluate \textbf{ExpertCondenser}, our proposed method, against two state-of-the-art MoE post-training approaches: ESFT~\citep{wang2024let} and DenseMixer~\citep{yao2025densemixer}. In~\S~\ref{subsection:expertcondenser_dissecting}, we present ablation studies showing that all components of our approach contribute to the overall performance improvement. Appendix~\ref{appendix:experiment_settings} further provides the detailed experimental settings. We provide additional analysis of cross-domain generalization in Appendix~\ref{appendix:ood_generalization} and training stability in Appendix~\ref{appendix:training_stability}.

\subsection{Math Reasoning}
\label{sec:experiment_math_reasoning}

We report performance on two math reasoning datasets (\textit{Math-7K}~\footnote{More details about the \texttt{Math-7K} dataset can be found in Appendix~\ref{appendix:math7k}.} and \textit{Math-14K}) in Table~\ref{tab:results_math7k_14k}. 
Across all benchmarks (GSM8K, SingleEq, SVAMP, MultiArith, AddSub, AQuA, and MAWPS), ExpertCondenser consistently outperforms both ESFT and DenseMixer on DeepSeek-V2-Lite, Qwen1.5-MoE, and OLMoE.
On \textit{Math-7K}, it improves the average accuracy of Qwen1.5-MoE by 3.3\%, DeepSeek-V2-Lite by 2.7\%, and OLMoE by 2.6\%, and GPT-OSS by 1.0\%. 
On \textit{Math-14K}, ExpertCondenser also boosts performance.  

Table~\ref{tab:dataset_results_stanfords1} further demonstrates that on most SoTA math reasoning benchmarks, ExpertCondenser outperforms baseline methods across Qwen3, DeepSeek-Coder-V2-Lite, and Qwen1.5~\footnote{OLMoE has a maximum context length of only 4096 tokens which is not suitable for \textit{Stanford-S1}.}. ExpertCondenser enhances accuracy of DenseMixer by 2.9\%, 1.7\%, and 2.0\% on Qwen3, DeepSeek-Coder-V2-Lite, and Qwen1.5 respectively.

\subsection{Other Datasets}
\label{sec:experiment_commonsense_reasoning}

To ensure that our findings above are generalizable, we further examine the performance of ExpertCondenser under the common sense reasoning dataset \texttt{CommonSense-15K}, including downstream test datasets, \texttt{BoolQ}, \texttt{PIQA}, \texttt{SIQA}, \texttt{HellaSwag}, \texttt{ARC-e}, \texttt{ARC-c}, and \texttt{OBQA}.  

Table ~\ref{tab:commonsense_results} reports the performance of DenseMixer, ESFT, and ExpertCondenser on the \texttt{CommenSense} dataset. We can observe that ExpertCondenser surpasses the DenseMixer by 2.2\% on post-trained OLMoE. On post-trained Qwen1.5-MoE, ExpertCondenser surpasses the best performance of DenseMixer and ESFT by 3.0\% and 3.9\%, respectively.

\begin{table*}[t!]
% \vspace{-0.3in}
\centering
\caption{Accuracy comparison of OLMoE and Qwen1.5-MoE with various post-training methods on commonsense
reasoning datasets.} 
\label{tab:commonsense_results}
\resizebox{\textwidth}{!}{
\setlength{\tabcolsep}{1.5mm}{
\begin{tabular}{p{2.5cm} c c c c c c c c c c c}
\toprule
\centering\textbf{Model} & \textbf{Model Size} & \textbf{Post-train method} & \textbf{BoolQ} & \textbf{PIQA} & \textbf{SIQA} & \textbf{HellaSwag} & \textbf{WinoGrande} & \textbf{ARC-e} & \textbf{ARC-c} & \textbf{OBQA} & \textbf{AVG} \\
\midrule

\multirow{4}{*}{\textbf{OLMoE}}  
& \multirow{4}{*}{\textbf{7B}} 
  & SFT          & 62.5 & 65.8 & 62.8 & 70.7 & 71.4 & 78.4 & 63.7 & 70.6 & 68.2 \\
&   & ESFT          & 63.5 & 63.0 & 58.9 & 64.7 & 62.8 & 74.8 & 63.8 & 63.4 & 64.4 \\
&   & DenseMixer    & 62.8 & 68.7 & 65.3 & 71.6 & 73.5 & 81.3 & 63.5 & 71.3 & 69.8 \\
&   & \textbf{ExpertCondenser (Ours)} & 67.7 & 71.4 & 69.7 & 69.8 & 75.8 & 76.8 & 71.0 & 73.6 & \textbf{72.0} \\

\midrule

\multirow{4}{*}{\textbf{Qwen1.5-MoE}}  
& \multirow{4}{*}{\textbf{14B}} 
  & SFT          & 68.8 & 84.7 & 74.5 & 76.8 & 75.6 & 84.6 & 72.8 & 76.4 & 76.8 \\
&   & ESFT          & 69.7 & 85.3 & 75.4 & 78.2 & 74.2 & 84.0 & 71.8 & 75.0 & 76.7 \\
&   & DenseMixer    & 70.8 & 85.7 & 74.6 & 75.8  & 78.9 & 82.6 & 74.8 & 77.8 & 77.6 \\
&   & \textbf{ExpertCondenser (Ours)} & 72.1 & 84.9 & 75.6 & 81.6 & 79.8 & 88.5 & 78.1 & 84.4 & \textbf{80.6} \\

\bottomrule
\end{tabular}
}}
\vspace{-0.1 in}
\end{table*}

\subsection{System Efficiency under Parameter Offloading and Expert Parallelism}
\label{sec:system_efficiency}

\begin{figure}[htbp]
\vspace{-0.4 in}
    \centering
    \begin{minipage}{0.50\textwidth}
        \centering
        \includegraphics[width=\textwidth]{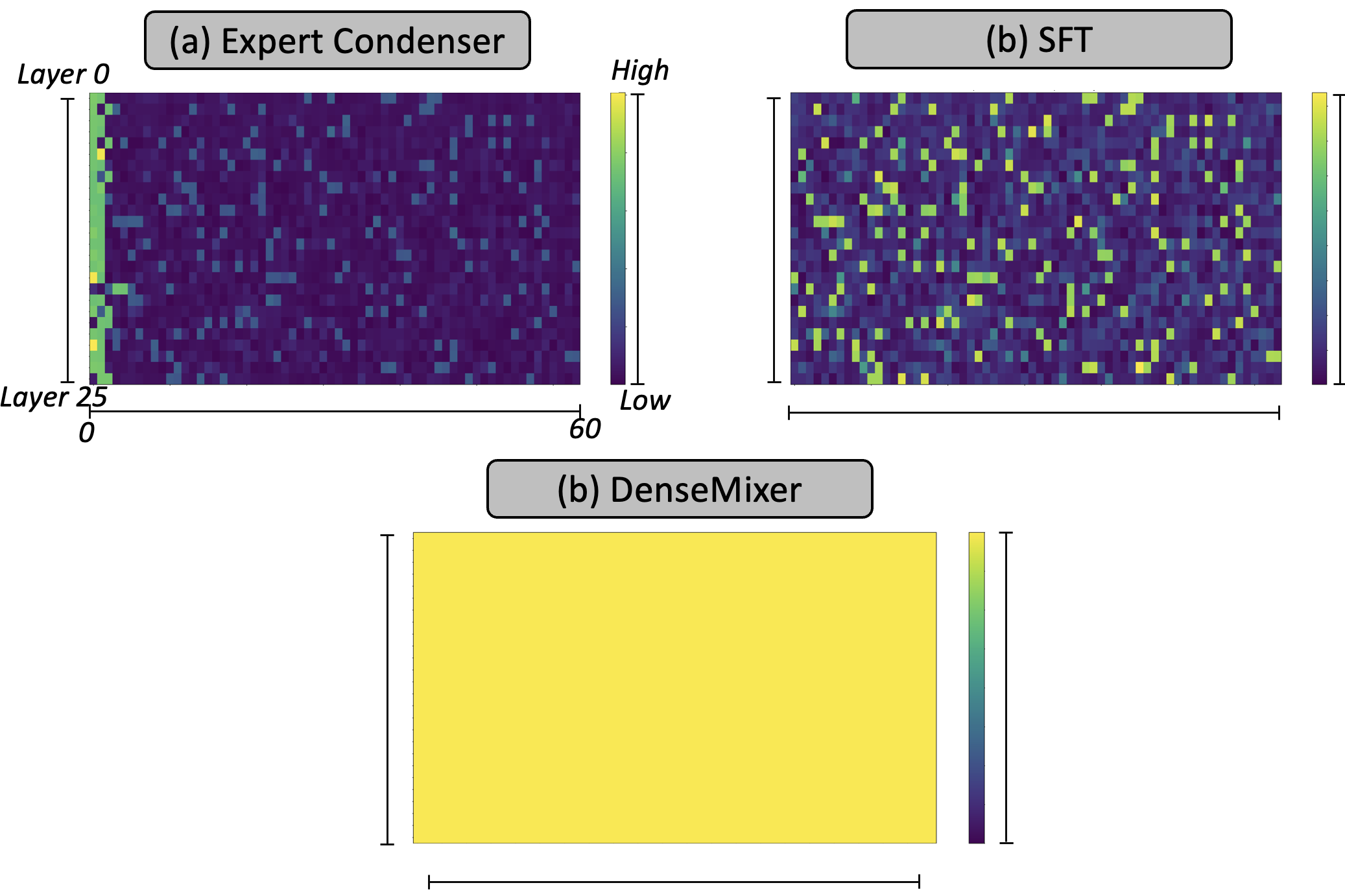}
        \vspace{-0.3in}
        \caption{Activation counts for all experts across layers for:(a) ExpertCondenser, (b) SFT, and (c) DenseMixer.}
        \vspace{-0.05in}
        
        \label{fig:expertcondenser_expert_rate}
    \end{minipage}
    \hfill
    \begin{minipage}{0.38\textwidth}
        \centering
        \captionof{table}{The experiments involved Post-training, ESFT, DenseMixer, and ExpertCondenser on 8 $\times$ H100 80GB GPUs using parameter offloading, with a batch size of 32. Communication between the GPU and CPU was facilitated via PCIe-G4.}
        \vspace{-0.05in}
        
        \label{tab:time-profiling-results}
        \adjustbox{max width=\textwidth}{
        \begin{tabular}{p{1.3cm} c c c c} 
            \toprule
            \multicolumn{4}{c}{\textbf{DeepSeek-V2-Lite}} \\
            \midrule
            \textbf{Methods} & \textbf{Activate \#Params\%} & \textbf{Time/s} & \textbf{Speedup} \\
            \midrule
            \textbf{DenseMixer}     & 16B  & 362.83 & 1$\times$ \\
            \textbf{SFT}            & 2.4B & 152.78 & 2.37$\times$ \\
            \textbf{ExpertCondenser} & 2.4B & 126.24 & 2.87$\times$ \\
            \bottomrule
        \end{tabular}
        }
    \end{minipage}
    \vspace{-0.15 in}
\end{figure}

Training large-scale MoE models under memory constraints often relies on parameter offloading (e.g., ZeRO-Offload~\citep{Ren2021ZeROOffloadDB}), where expert weights are dynamically transferred between CPU and GPU. In such settings, efficiency is sensitive to expert activation patterns, as frequent parameter movement can become a major bottleneck.

Although our primary goal is to improve post-training effectiveness, ExpertCondenser naturally exhibits a favorable systems property under offloading. Condenser experts are always selected across tokens, resulting in consistently high activation frequency. This allows their parameters to remain resident in GPU memory, reducing CPU--GPU transfers.

Figure~\ref{fig:expertcondenser_expert_rate}(a) shows that Condenser experts dominate activations across layers, and therefore never need to be swapped out of GPU memory. In contrast, SFT selects Top-$k$ experts dynamically, leading to volatile activation patterns (Figure~\ref{fig:expertcondenser_expert_rate}(b)) and increased offloading overhead. DenseMixer activates all experts (Figure~\ref{fig:expertcondenser_expert_rate}(c)), removing sparsity benefits and requiring all experts to be loaded into GPU memory.
We evaluate this effect under parameter offloading on DeepSeek-V2-Lite (Table~\ref{tab:time-profiling-results}). ExpertCondenser achieves a $2.87\times$ speedup over DenseMixer and outperforms SFT.

Offloading is a single-node, memory-constrained regime. Large MoE models are more commonly served under \emph{expert parallelism} (EP), where experts are sharded across devices and every routed expert computation crosses an all-to-all. 
To exam directly under EP determines whether the condenser experts design helps or hurts, we replicate condenser experts on every rank, exactly as Qwen1.5-MoE, DeepSeek-V2-Lite, OLMoE, and GPT-OSS already replicate their shared experts under EP. Replication costs $r/n$ of the MoE parameters per device ($3.3\%$ at $r{=}2$ against the $n{=}60$ routed experts of Qwen1.5-MoE) and in exchange keeps $r$ of each token's $k$ expert computations on-device, so all-to-all volume falls by a factor of $r/k$. Table~\ref{tab:ep-benchmark} confirms both effects on $8\times$H100. ExpertCondenser moves half the dispatch bytes of SFT ($0.500\times$, matching the predicted $r/k = 2/4$ for Qwen1.5-MoE) and completes a step in $0.800\times$ its time, while DenseMixer pays $16.0\times$ the traffic. Peak per-rank memory also falls below SFT ($0.66$ vs.\ $0.84$\,GB), because the smaller all-to-all buffers outweigh the two replicated experts. The always-active pattern that helps under offloading is therefore neutral-to-favorable for EP communication as well.

We emphasize that this efficiency gain is a secondary effect of the proposed design. Nonetheless, it highlights that stable expert activation can improve training efficiency in both offloading- and EP-based settings.

\begin{table}[t]
\centering
\caption{Expert-parallel benchmark on $8\times$H100 with one Qwen1.5-MoE-A2.7B MoE layer.\protect\footnotemark}
\label{tab:ep-benchmark}
\resizebox{0.8\textwidth}{!}{%
\vspace{-0.05in}
\begin{tabular}{l c c c}
\toprule
\textbf{Method} & \textbf{Step time (ms)} & \textbf{Dispatch (MB/step)} & \textbf{Peak mem/rank (GB)} \\
\midrule
DenseMixer      & 67.00 ($3.109\times$) & 7516.19 ($16.017\times$) & 10.00 \\
SFT             & 21.55 ($1.000\times$) & 469.26 ($1.000\times$)   & 0.84 \\
\textbf{ExpertCondenser} & \textbf{17.24} ($0.800\times$) & \textbf{234.68} ($0.500\times$) & \textbf{0.66} \\
\bottomrule
\end{tabular}
}
\vspace{-0.12in}
\end{table}
\footnotetext{$32{,}768$ tokens per step, bf16, NCCL \texttt{all\_to\_all\_single}, 5 warmup and 20 timed steps. Step time is the median forward\,+\,backward time max-reduced across ranks; dispatch bytes are summed over the 8 ranks. The $r{=}2$ condenser experts are replicated on every rank.}

\subsection{Dissecting the ExpertCondenser Algorithm}
\label{subsection:expertcondenser_dissecting}

Table~\ref{tab:mix_dataset_results} presents an ablation study of ExpertCondenser on DeepSeek-V2-Lite and Qwen1.5-MoE under the \textbf{math7k} benchmark. 
We evaluate progressively variants: 
(1) \emph{aux-free}, removing auxiliary losses; 
(2) \emph{aux-free+bias}, introducing bias-based routing; 
(3) \emph{aux-free+bias+condenser experts}, enabling always-active gated experts; and 
(4) \emph{with-aux+condenser experts}, which retains auxiliary loss while adding condenser experts. 

Results from (1) to (3) show that each component contributes incrementally, while their combination yields the best performance.
Results in (4) further reveal that incorporating condenser experts on top of the auxiliary-loss setting still improves performance over the SFT baseline, demonstrating that the benefit of condenser experts.

A further control isolates the gating itself: replacing the router weight $g_j(x_t)$ on the condenser path with a fixed ungated shared expert removes the gain, so the benefit comes from the persistent path being token-dependent rather than merely always active (Appendix~\ref{appendix:gating_ablation}).

\begin{table*}[h!]
\centering
\caption{Evaluation of variant post-trained models on downstream Math Reasoning tasks.}
\label{tab:mix_dataset_results}
\vspace{-0.1in}
\resizebox{\textwidth}{!}{%
  \setlength{\tabcolsep}{1.5mm}%
  \begin{tabular}{p{1.5cm} p{2.8cm} c c c c c c c c c}
    \toprule
    \multirow{2}{*}{\textbf{Dataset}} 
      & \multirow{2}{*}{\textbf{\centering Model}} 
      & \multirow{2}{*}{\textbf{Post-train Type}}
      & \multirow{2}{*}{\textbf{GSM8k}}
      & \multirow{2}{*}{\textbf{SingleEq}}
      & \multirow{2}{*}{\textbf{SVAMP}} 
      & \multirow{2}{*}{\textbf{MultiArith}} 
      & \multirow{2}{*}{\textbf{AddSub}} 
      & \multirow{2}{*}{\textbf{AQuA}} 
      & \multirow{2}{*}{\textbf{mawps}} 
      & \multirow{2}{*}{\textbf{AVG}} \\
    & & & & & & & & & & \\
    
    \midrule
    \multirow{8}{*}{\textbf{math7k}} 
      & \multirow{4}{*}{\textbf{DeepSeek-V2-Lite}}
      & aux-free+bias+condenser experts & 59.4 & 92.5 & 69.1 & 91.5 & 79.5 & 36.1 & 83.6 & 73.1 \\
      & & aux-free+bias & 58.8 & 90.7 & 69.3 & 88.7 & 74.2 & 36.1 & 80.3 & 71.2 \\
      & & aux-free & 57.6 & 89.6 & 68.6 & 87.5 & 73.8 & 35.2 & 80.4 & 70.4 \\
      & & with-aux+condenser experts & 58.3 & 88.4 & 68.9 & 88.6 & 75.8 & 35.6 & 81.7 & 71.0 \\

    \cmidrule(l){2-11}
    & \multirow{4}{*}{\textbf{Qwen1.5-MoE}} 
      & aux-free+bias+condenser experts & 57.2 & 74.6 & 55.7 & 86.0 & 61.8 & 33.1 & 75.6 & 63.4 \\
      & & aux-free+bias & 48.2 & 76.6 & 52.7 & 80.3 & 59.2 & 26.8 & 71.8 & 59.4 \\
      & & aux-free & 47.2 & 74.0 & 51.8 & 82.0 & 58.7 & 30.3 & 71.8 & 59.0 \\
      & & with-aux+condenser experts & 49.5 & 76.8 & 53.6 & 82.7 & 60.1 & 30.9 & 72.6 & 60.9 \\
    \bottomrule
  \end{tabular}%
}

\vspace{-0.12in}
\end{table*}

\section{Further Analysis}

% \vspace{-0.12in}
\subsection{Expert Routing and Output Analysis}
\label{subsection:routing_and_output_analysis}

\begin{figure}[h]
\centering

% ===== LEFT: TABLE =====
\begin{minipage}{0.52\textwidth}
    \centering
    \vspace{-0.05in}
    \centering
    \includegraphics[width=0.9\linewidth]{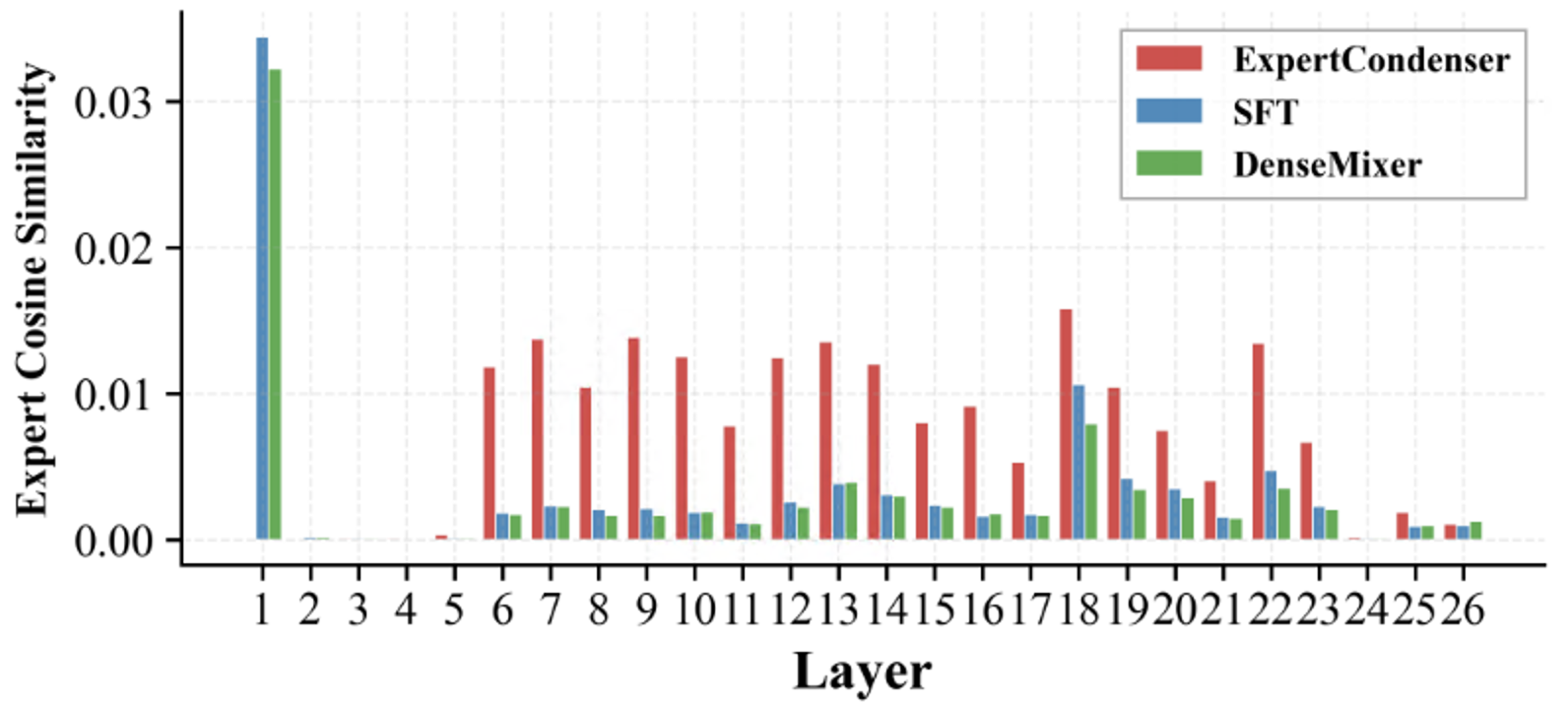}
    % \caption{Average FFN gradient norm per expert during training. Condenser Experts (blue) maintain consistently higher and more stable gradient magnitudes, while routed experts (orange) exhibit lower and more sporadic gradients due to routing sparsity.}
    \vspace{-0.1in}
    \caption{Weighted MoE Output Divergence.}
    \label{fig:moe_output_divergence}

\end{minipage}
\hfill
% ===== RIGHT: FIGURE =====
\begin{minipage}{0.46 \textwidth}
    \centering
    \vspace{-0.05in}
    \includegraphics[width=0.95\textwidth]{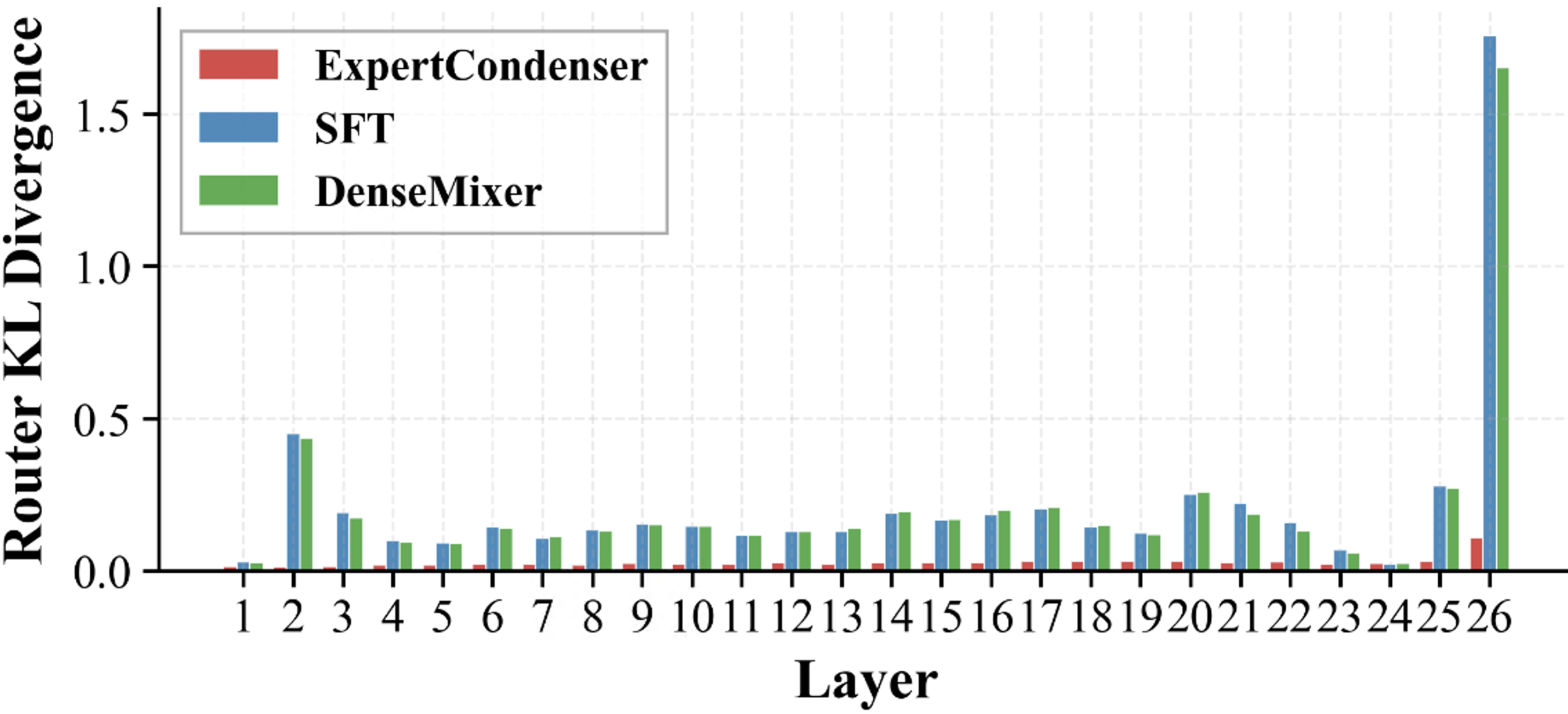}
    \caption{Router KL Divergence.}
    \label{fig:router_kl_divergence}
        
\end{minipage}
\vspace{-0.12in}
\end{figure}

To understand how different methods adapt MoE models, we analyze two metrics:
(1) \textbf{Router KL Divergence} in Fig.~\ref{fig:moe_output_divergence}, which measures shifts in routing distributions relative to the base model, and 
(2) \textbf{Weighted MoE Output Divergence} in Fig.~\ref{fig:router_kl_divergence}, which measures changes in layer outputs under each model’s routing weights.

As shown in Fig.\ref{fig:router_kl_divergence}, both SFT and DenseMixer significantly alter routing distributions, indicating that adaptation is partly achieved by rerouting tokens to different experts. In contrast, ExpertCondenser exhibits minimal router divergence, preserving the base model's routing structure. Meanwhile, Fig.\ref{fig:moe_output_divergence} reveals that ExpertCondenser induces more substantial changes in expert outputs than other methods. Together, these results suggest that ExpertCondenser adapts MoE models by directly transforming expert parameters rather than disrupting fragile routing behavior, supporting our motivation in~\S~\ref{subsection:Auxiliary-Free-Sparsity-Enforcement}. The forced expert mechanism and learned biases enable this targeted adaptation, maintaining training-inference alignment while concentrating representational changes within the experts themselves.
Preserving the base router also shows up downstream: after fine-tuning on general instruction data that contains no mathematical supervision, ExpertCondenser retains more of the base model's math ability than SFT (Appendix~\ref{appendix:ood_generalization}).

\subsection{Gradient Norm Analysis}
\label{subsection:gradient_norm_analysis}

To empirically validate the gradient starvation effect and the role of Condenser Experts, we analyze the gradient norms of expert FFN parameters during training. Specifically, for each expert, we measure the $\ell_2$ norm of its parameter gradients and report the average gradient norm across (i) Condenser Experts and (ii) standard routed experts at each training step.

\begin{figure}[h]
\centering

% ===== LEFT: TABLE =====
\begin{minipage}{0.58\textwidth}
    \vspace{-0.2in}
    \centering
    \label{tab:expert_condense_analysis}
    \centering
    \includegraphics[width=0.9\linewidth]{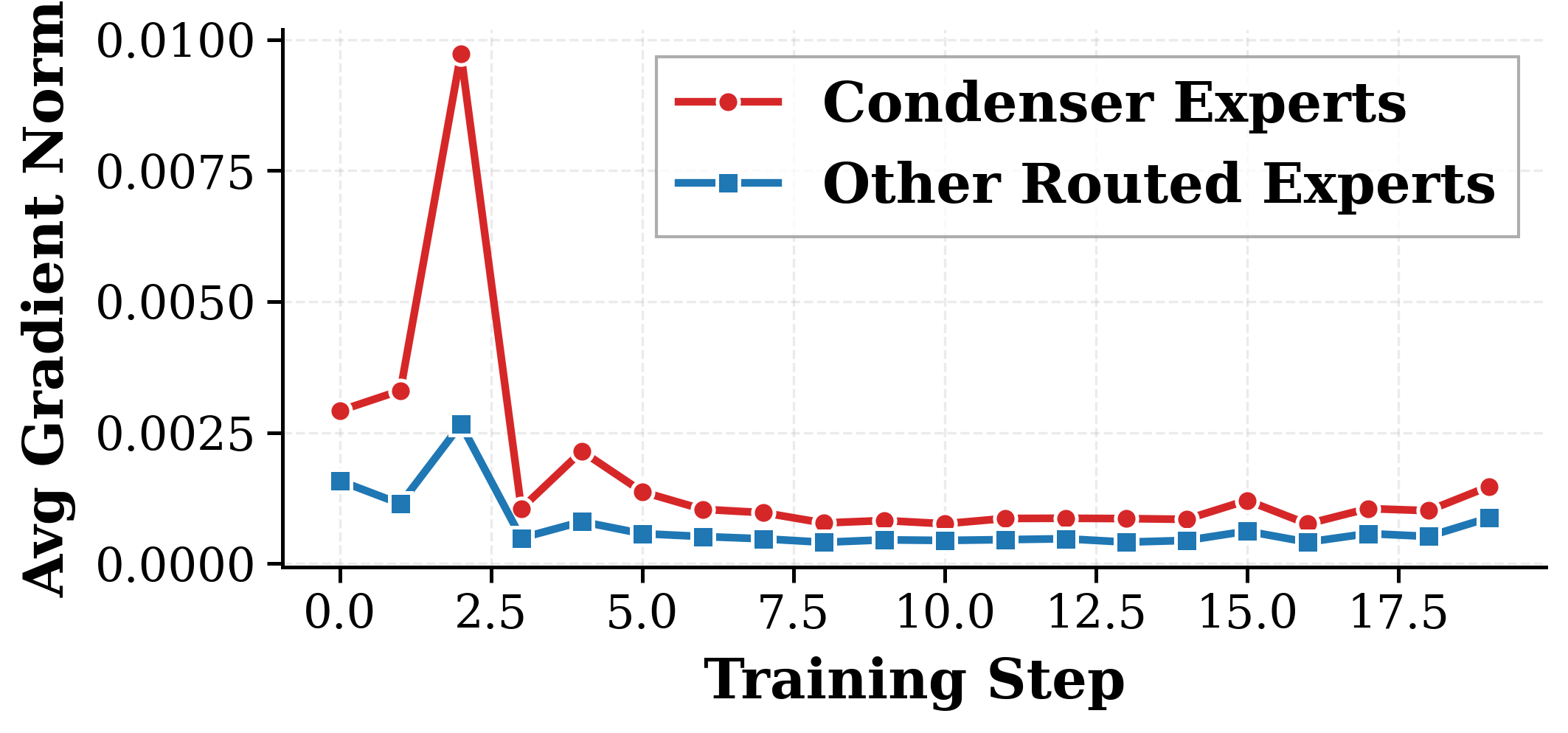}
    % \caption{Average FFN gradient norm per expert during training. Condenser Experts (blue) maintain consistently higher and more stable gradient magnitudes, while routed experts (orange) exhibit lower and more sporadic gradients due to routing sparsity.}
    \vspace{-0.13in}
    \caption{Average FFN gradient norm per expert during training.}
    \label{fig:gradient_norm}

\end{minipage}
\hfill
% ===== RIGHT: FIGURE =====
\begin{minipage}{0.38\textwidth}
    \centering
    \includegraphics[width=\textwidth]{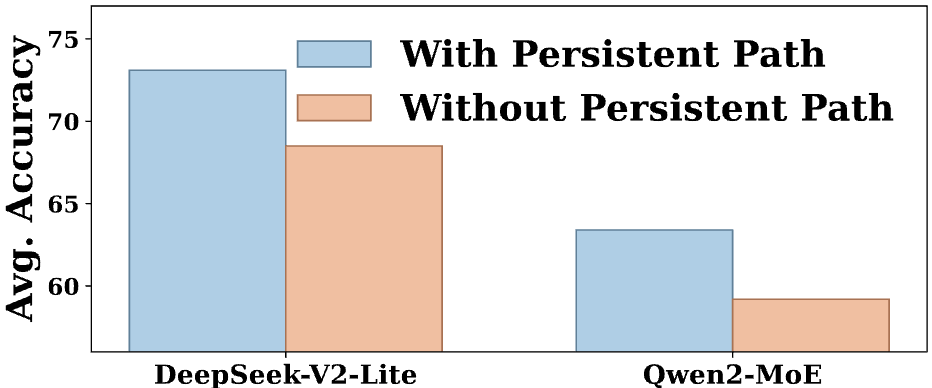}
    \vspace{-0.23in}
    \caption{Persistent path ablation. “With” uses always-active gated condenser experts; “w.o.” treats them as routed experts (i.e., not always selected) during inference.}
    \label{fig:persistent_path}
\end{minipage}
\vspace{-0.3in}
\end{figure}

Figure~\ref{fig:gradient_norm} shows the evolution of gradient norms over training. We observe that routed experts exhibit lower gradient magnitudes, consistent with their limited activation probability. In contrast, condenser experts maintain consistently higher gradient norms throughout training. These results provide empirical support for our theory in ~\S\ref{assumption:avoid_gradient_bottleneck}, suggesting that condenser experts effectively mitigate gradient starvation and enable sustained learning across training, and persistent gradient flow is crucial to retain task-relevant information.

\subsection{Condenser Experts Analysis}
\label{subsection:condenser_experts_analysis}

We investigate the following research questions in this subsection:
\textbf{Do condenser experts facilitate information consolidation from long-tailed experts?} To answer this, we examine two empirical signatures: Persistent path ablation in this subsection and expert correlation analysis in Appendix~\ref{appendix:expert_correlation_analysis}.
\textbf{For persistent path ablation.} We test the consolidation hypothesis in~\S~\ref{assumption:consolidate_info} via a causal intervention on the persistent path. After training, we compare standard inference—where condenser experts are always selected—with a variant in which they are treated as standard routed experts. This preserves model capacity, parameters, and activation budget, while removing only the persistent aggregation pathway.

If condenser experts do not accumulate task-relevant information, this change should have minimal effect; otherwise, performance should degrade. As shown in Fig.~\ref{fig:persistent_path}, we observe consistent drops on Math7K for both Qwen1.5-MoE and DeepSeek-V2-Lite, indicating that condenser experts indeed consolidate task-relevant information. The observation further provides empirical support for our theoretical analysis in ~\S~\ref{assumption:consolidate_info}.

\section{Conclusion}
\vspace{-0.1in}

Our empirical findings reveal that even rarely activated experts encode indispensable knowledge, and pruning them directly leads to substantial performance degradation. 
Motivated by this observation, we proposed ExpertCondenser, a post-training framework that combines bias-driven sparsification with consistently active gated experts. This design alleviates gradient starvation, narrows the train–inference routing gap, and provides a persistent pathway that helps preserve and consolidate information. Across math and commonsense reasoning benchmarks, ExpertCondenser consistently outperforms existing post-training methods such as ESFT and DenseMixer.

\section*{Acknowledgments}

We extend our gratitude to Yu (Ivy) Yang for her helpful suggestions on manuscript preparation and research presentation.

This research was conducted using the ORCHARD high-performance computing cluster operated by the Office of Research Computing and Data at Carnegie Mellon University. We gratefully acknowledge Carnegie Mellon University for providing access to these shared computing resources.

This research also used the Bridges-2 system at the Pittsburgh Supercomputing Center and the Delta advanced computing system. These resources are supported by the National Science Foundation under Grants No. 2138259, 2138286, 2138307, 2137603, and 2138296.

\bibliography{colm2026_conference}
\bibliographystyle{colm2026_conference}

%%%%%%%%%%%%%%%%%%%%%%%%%%%%%%%%%%%%%%%%%%%%%%%%%%%%%%%%%%%%%%%%%%%%%%%%%%%%%%%
%%%%%%%%%%%%%%%%%%%%%%%%%%%%%%%%%%%%%%%%%%%%%%%%%%%%%%%%%%%%%%%%%%%%%%%%%%%%%%%
% APPENDIX
%%%%%%%%%%%%%%%%%%%%%%%%%%%%%%%%%%%%%%%%%%%%%%%%%%%%%%%%%%%%%%%%%%%%%%%%%%%%%%%
%%%%%%%%%%%%%%%%%%%%%%%%%%%%%%%%%%%%%%%%%%%%%%%%%%%%%%%%%%%%%%%%%%%%%%%%%%%%%%%
\newpage
\appendix

\newpage
\clearpage

\section{Notation}
\label{sec:notation}

For clarity, we summarize the main notations used throughout the theoretical sections.

\begin{itemize}
    \item[$h_i$] Raw output vector of expert $i$.
    \item[$v_i$] Decomposed expert output: $v_i = \rho_i e_i$.
    \item[$\rho_i$] Magnitude (norm) of expert $i$’s output.
    \item[$e_i$] Normalized direction vector of expert $i$, $\|e_i\|=1$.
    \item[$y$] Dense output: $y=\sum_{i=1}^n v_i$.
    \item[$\hat{y}$] Sparse approximation using only $k$ experts.
    \item[$S$] Index set of selected experts, $|S|=k$.
    \item[$\lambda_i$] Binary indicator of expert selection.
    \item[$n,k$] Total number of experts, and number of selected experts.
    \item[$v^{(l)}_{i,j,k}$] Output of expert $i$ at layer $l$ for token $k$ in $x_j$.
    \item[$g^{(l)}_{i,j,k}$] Gate score assigned to expert $i$ at layer $l$ for token $k$.
    \item[$g_{i,t}$] Indicator if expert $i$ is selected for token $t$ (shorthand).
    \item[$s_{i,t}$] Gating score (logit) of expert $i$ for token $t$ before normalization.
    \item[$w_{i,t}$] Softmax-normalized routing weight assigned to expert $i$ for token $t$, and softmax-normalized routing weight from the \emph{unbiased} scores $s_{i,t}$.
    \item[$L_j$] Length (number of tokens) of sample $x_j$.
    \item[$\mathcal{D}_s, N_s$] Subset of training data and its size.
    \item[$x_j$] $j$th token for training
    \item[$s^{(l)}_i$] Magnitude-based expert score (ES-Mag) at layer $l$.
    \item[$r^{(l)}_i$] Activation-ratio expert score (ES-Act) at layer $l$.
    \item[$K$] Number of experts selected per token.
    \item[$S^{(l)}$] Set of selected experts at layer $l$.
    \item[$\mathcal{L}_{\text{Balance}}$] Auxiliary loss for balancing expert utilization.
    \item[$f_i$] Normalized fraction of tokens routed to expert $i$.
    \item[$P_i$] Average routing weight assigned to expert $i$ across tokens.
    \item[$\alpha$] Hyperparameter controlling auxiliary loss strength.
    \item[$T$] Sequence length (number of tokens).
    \item[$\theta$] Router parameters used to compute gating scores.
    \item[$\mathcal{L}$] Generic training loss depending on model output $y$.
    \item[$b_i$] Expert-wise bias used only for selection to improve load balance.
    \item[$\tilde{s}_{i,t}$] Biased gating score for expert $i$ on token $t$: $\tilde{s}_{i,t}=s_{i,t}+b_i$.
    \item[$S_t$] Top-$K$ selection set for token $t$ obtained from $\{\tilde{s}_{j,t}\}_{j=1}^n$.
    \item[$\gamma$] Bias update speed.
\end{itemize}

\clearpage
\newpage

\section{Learnable Persistent Mixing Principle}
\label{appendix: LPM}
Modern large-scale architectures increasingly rely on multi-path computation.
Similar to Hyper-Connection, the designing of Mixture-of-Experts (MoE) share the same philosophy: increase MLP layer outputs’ topological by composing multiple parallel experts or computational paths. 
However, beyond structural similarity, HC reveals a deeper principle which we named \textbf{Learnable Persistent Mixing(LPM): persistent computational paths should be governed by learnable mixing operators rather than fixed injection rules}.
In this appendix section, we formalize this principle and show how it theoretically motivates our Condenser expert design.
\paragraph{Learnable Persistent Mixing}
Consider a collection of $n$ parallel candidate transformations 
$\{f_i(\cdot)\}_{i=1}^n$, each mapping $\mathbb{R}^d \rightarrow \mathbb{R}^d$.
A general multi-path layer can be written as
\begin{equation}
\mathbf{h}'(\mathbf{x})
=
\sum_{i=1}^{n}
m_i(\mathbf{x})\, f_i(\mathbf{x}),
\qquad
m(\mathbf{x}) \in \mathcal{M} \subseteq \mathbb{R}^n,
\label{eq:general_mixing}
\end{equation}
where $m(\mathbf{x})$ defines how information flows across parallel paths.
This decomposition separates computation (the $f_i$) from connectivity (the mixing coefficients $m_i$).
Different architectures correspond to different parameterizations and constraints on the admissible mixing set $\mathcal{M}$.

\paragraph{Hyper-Connection Serve as an Example of LPM}

Hyper-Connections~\citep{zhu2024hyper, xie2025mhc} explicitly parameterize connectivity across parallel residual streams.
Let $\mathbf{x}_\ell \in \mathbb{R}^{n \times d}$ denote the expanded residual representation at layer $\ell$.
The HC update can be written in stream-wise form as
\begin{equation}
\mathbf{x}_{\ell+1,j}
=
\sum_{i=1}^{n}
H^{\mathrm{res}}_{\ell,ji}\,\mathbf{x}_{\ell,i}
+
H^{\mathrm{post}}_{\ell,j}
F_\ell
\!\left(
\sum_{i=1}^{n}
H^{\mathrm{pre}}_{\ell,i}\,\mathbf{x}_{\ell,i}
\right),
\label{eq:hc_update}
\end{equation}
where $H^{\mathrm{res}}_\ell \in \mathbb{R}^{n\times n}$,
$H^{\mathrm{pre}}_\ell \in \mathbb{R}^{1 \times n}$,
and $H^{\mathrm{post}}_\ell \in \mathbb{R}^{1 \times n}$ are learnable parameters.
Equation~\eqref{eq:hc_update} reveals that every persistent path in HC is modulated by learnable coefficients.
Residual propagation is controlled by $H^{\mathrm{res}}_\ell$, rather than being a fixed identity map.
The transformed branch is read and written via $H^{\mathrm{pre}}_\ell$ and $H^{\mathrm{post}}_\ell$.
This full parameterization of connectivity is central to HC's expressive power. \textbf{Hyper-Connections provide a useful lens: persistent paths are most expressive when they remain learnably mixed rather than fixed.}

\paragraph{Condenser Experts Instead of Static Shared Experts} Traditional shared component 
\[
h_{\text{static}}(x)=\sum_{i=1}^{\hat{n}} E_i(x)
\]
is injected without any mixing weight. It is always active and contributes with fixed unit weight. Motivated by the Learnable Persistent Mixing (LPM) principle revealed by Hyper-Connections, we argue that persistent shared computation in MoE should also be governed by learnable mixing

\[
h_{cond} =  
\sum_{j \in J} g_j(x_t)\, E_j(x_t)
\]

Condenser experts serve as a special case of persistent computation paths, where a subset of experts is always active across tokens. Unlike static residual pathways, these experts remain input-gated, aligning with LPM and prior observations~\citep{xie2025mhc, zhu2024hyper} that persistent paths benefit from adaptive mixing. This interpretation complements our primary motivation of preserving long-tailed expert knowledge.

\clearpage
\newpage

\section{Experimental Settings}
\label{appendix:experiment_settings}

\textbf{Model Architecture and Dataset:} In our experimental setup, we use open-weight GPT-OSS-20B~\citep{openai2025gptoss120bgptoss20bmodel}, Deepseek-V2-Lite~\citep{liu2024deepseek}, Deepseek-Coder-V2-Lite-Instruct~\citep{liu2024deepseek}, OLMoE-7B-01-25~\citep{muennighoff2024olmoeopenmixtureofexpertslanguage}, Qwen1.5-MoE-A2.7B~\citep{qwen2}, and Qwen3-30B-A3B~\citep{qwen3} to conduct experiments. 

In Section\S~\ref{sec:scaling_laws}, We study the MoE expert scaling law using Deepseek-Coder-V2-Lite-Instruct base model and GPT-OSS-20B base model, and test the evaluation performance on \texttt{MultiArith}, \texttt{GSM\_8K}~\citep{cobbe2021gsm8k}, \texttt{AddSub}, 
\texttt{AQuA}, \texttt{SingleEq}, \texttt{SVAMP}, and \texttt{mawps}.

In Subsection \S~\ref{sec:experiment_math_reasoning}, we evaluate MoE post-training algorithms on math reasoning domains. We fine-tune on the \texttt{Math7K} and \texttt{Math14K} dataset using the DeepSeek-V2-Lite, Qwen1.5-MoE-A2.7B, OLMoE-7B-01-25, and GPT-OSS-20B and test the evaluation performance on downstream testsets \texttt{MultiArith}, \texttt{GSM\_8K}~\citep{cobbe2021gsm8k}, \texttt{AddSub}, 
\texttt{AQuA}, \texttt{SingleEq}, \texttt{SVAMP}, and \texttt{mawps}. Then, we post-train Qwen1.5-MoE-A2.7B, DeepSeek-Coder-V2-Lite-Instruct, and Qwen3-30B-A3B on \texttt{Stanford-S1} dataset~\citep{muennighoff2025s1simpletesttimescaling} and test the evaluation performance on downstream SOTA math reasoning benchmarks \texttt{AIME2025}, \texttt{AIME2024}, \texttt{GPQA-Diamond}~\citep{rein2024gpqa}, and \texttt{MATH-500}.

In Subsection \S~\ref{sec:experiment_commonsense_reasoning}, we turn to commonsense reasoning. Following~\citep{hu2023llm, he2024sparse, liu2024dora}, we merge the training sets of eight tasks into~\texttt{commonsense\_15k} and evaluate on their individual test sets: \texttt{BoolQ}, \texttt{PIQA}, \texttt{SIQA}, \texttt{HellaSwag}, \texttt{ARC-e}, \texttt{ARC-c}, and \texttt{OBQA}. Results are reported as accuracy, with an averaged score summarizing overall effectiveness. Across all datasets—\texttt{Stanford-S1K}, \texttt{Commonsense}, \texttt{Math7K}, and \texttt{Math14K}—our setup emphasizes the generalization ability of LLMs across diverse sub-tasks. 

\textbf{Training Framework and Hyper-parameters:} We used the \texttt{huggingface-trl}~\citep{vonwerra2022trl} library with zero-2 or zero-3~\citep{Ren2021ZeROOffloadDB} for fine-tuning and \texttt{vllm}~\citep{kwon2023efficient}, \texttt{lighteval}~\citep{lighteval}, and \texttt{accelerate}~\citep{accelerate} library for inference evaluation. Both training and evaluation are using dtype BF16. 

\textbf{MoE Post-train Baselines:} For state-of-the-art (SOTA) MoE post-training baselines, we choose to include ESFT~\citep{hu2021lora} and  DenseMixer~\citep{yao2025densemixer}. We re-implemented ESFT and DenseMixer following the setups in~\citep{yao2025densemixer}.
% The number of epochs, learning rate, and batch size will be the same when conducting experiment on the same model to ensure a fair comparison across different methods. 
% % For each method, we conduct a grid search on the training hyper-parameters (learning rate, batch size, etc) and report the best performance.

\textbf{Computational Resources:} We conduct our experiments and implement SOTA baselines of ESFT and DenseMixer~\cite{yao2025densemixer} to post-train with 8 NVIDIA H100\_80GB GPUs. Communication between the CPU and GPU is facilitated via PCIe-G4 and communication between GPUs is facilitated via Nvlink-3.

\clearpage
\newpage

\paragraph{Baseline Reimplementation and Fair Comparison.}

To ensure a fair comparison across methods, we adopt a unified training protocol. Some baseline results are not directly available in prior work under our exact model and dataset configurations. Therefore, we re-implement and fine-tune all baselines within a unified training framework. All methods are trained with identical learning rates, batch sizes, optimizers, and number of epochs on the same model. For method-specific components (e.g., Number of experts trained in ESFT), we follow the configurations recommended in their original works. All models are initialized from the same pretrained checkpoints and trained on identical datasets and splits. We also repeat experiments with multiple random seeds and observe consistent performance gains of ExpertCondenser, with improvements exceeding variance across runs.

As an example, for the \textbf{Math7K} dataset with \textbf{DeepSeek-V2-Lite}, the following configuration is used consistently across all methods:

\begin{table}[h]
\centering
\small
\begin{tabular}{l|l}
\hline
\textbf{Parameter} & \textbf{Value} \\
\hline
Data Type & bf16 \\
Optimizer & AdamW \\
Learning Rate & $1 \times 10^{-5}$ \\
Batch Size & 32 \\
Sequence Length & 4096 \\
Seed & 1234 \\
Warmup Ratio & 0.1 \\
LR Scheduler & cosine\_with\_min\_lr \\
Attention Implementation & flash\_attention\_2 \\
ZeRO Stage & 3 \\
\hline
\end{tabular}
\caption{Example training configuration used across all methods for Math7K with DeepSeek-V2-Lite.}
\label{tab:training_config}
\end{table}

All baselines and our method are trained with the above configuration. More hyper-parameters settings can also be found in our open-sourced repository.

\newpage
\clearpage

\section{Table Results for Experts Scaling Laws}
\label{appendix:scale_law_tables}

Tables~\ref{tab:scale_laws_gpt} and \ref{tab:scale_laws_deepseek} report detailed results of our scaling law experiments on two representative MoE models: GPT-OSS-20B and DeepSeek-Coder-V2-Lite. 
We evaluate performance under three pruning strategies introduced in Section~\S\ref{sec:scaling_laws}:  

(1) \emph{Small Dense Conversion}, where the number of experts is reduced from $n$ to $n'$ and all surviving experts are activated ($k=n'$);  

(2) \emph{Inference Reduction}, where the total number of experts is fixed ($n'=n$) but the activation budget is reduced from $k$ to $k' < k$; and  

(3) \emph{Small MoE Conversion}, where both the total number of experts is reduced ($n'<n$) and the activation budget is kept sparse ($k<n'$).  

% TABLE 2 - GPT-OSS experiments

\begin{table*}[htbp]
\centering
\caption{Evaluating base GPT-OSS-20B model Zero-Shot Results on downstream  Math Reasoning dataset, includes SingleEQ, MultiArith, AddSub, GSM8K, SVAMP, and AQuA.}
\label{tab:scale_laws_gpt}
\vspace{-0.1in}
\resizebox{\textwidth}{!}{%
  \setlength{\tabcolsep}{1.2mm}% tighter columns for metrics
  \begin{tabular}{
    p{3.8cm}  % Model (wide, single multirow cell)
    p{3.0cm}  % Strategies (wider for readability)
    c         % Remain Experts (n')
    c         % Activate Experts (k')
    c c c c c c c c  % metrics
  }
    \toprule
    \textbf{Model} & \textbf{Strategies} & \textbf{Remain Experts ($n'$)} & \textbf{Activate Experts ($k'$)} &
    \textbf{GSM8k} & \textbf{SingleEq} & \textbf{SVAMP} & \textbf{MultiArith} &
    \textbf{AddSub} & \textbf{AQuA} & \textbf{mawps} & \textbf{AVG} \\
    \midrule

    %================== One big multirow for the Model column (17 rows total) ==================%
    \multirow{17}{*}{\textbf{GPT-OSS-20B}} 
      & Base Model        & \textbf{32} & \textbf{4}  & 78.0 & 84.6 & 84.1 & 92.0 & 81.0 & 33.9 & 92.0 & 77.9 \\
    \cmidrule(l){2-12}
      & \multirow{6}{*}{Small Dense}
                           & \textbf{8}  & \textbf{8}   & 2.3 & 1.7 & 1.8 & 2.7 & 1.8 & 16.7 & 2.3 & 4.2 \\
      &                    & \textbf{12} & \textbf{12}  & 4.8 & 17.8 & 14.6 & 18.4 & 18.6 & 22.8 & 17.3 & 16.3 \\
      &                    & \textbf{18} & \textbf{18}  & 48.6 & 39.6 & 47.8 & 52.6 & 63.7 & 23.8 & 47.8 & 46.3 \\
      &                    & \textbf{24} & \textbf{24}  & 56.9 & 58.2 & 63.5 & 68.5 & 67.5 & 24.6 & 58.7 & 56.8 \\
    \cmidrule(l){2-12}
      & \multirow{3}{*}{Inference Reduce}
                           & \textbf{32} & \textbf{1}  & 5.5 & 19.7 & 13.6 & 12.8 & 19.5 & 23.2 & 18.1 & 16.1 \\
      &                    & \textbf{32} & \textbf{2}  & 70.8 & 74.6 & 76.4 & 88.2 & 75.4 & 29.9 & 76.9 & 70.3 \\
      &                    & \textbf{32} & \textbf{3} & 75.1 & 83.9 & 84.3 & 93.3 & 81.3 & 33.1 & 80.7 & 76.0 \\
      % &                    & \textbf{32} & \textbf{4}   & 78.4 & 84.8 & 84.1 & 91.3 & 77.7 & 34.6 & 84.5 & 76.5 \\
    \cmidrule(l){2-12}
      & \multirow{5}{*}{Small MoE}
                           & \textbf{8} & \textbf{4} & 1.6 & 2.2 & 3.2 & 3.2 & 2.0 & 12.6 & 2.5 & 3.9 \\
      &                    & \textbf{12} & \textbf{4}  & 6.4 & 18.7 & 21.8 & 13.0 & 18.2 & 21.7 & 16.0 & 16.5 \\
      &                    & \textbf{16} & \textbf{4}   & 20.2 & 43.7 & 48.1 & 43.7 & 41.0 & 24.0 & 42.4 & 37.6 \\
      &                    & \textbf{20} & \textbf{4}   & 32.8 & 56.1 & 59.9 & 58.2 & 52.9 & 25.6 & 55.5 & 48.7 \\
      &                    & \textbf{24} & \textbf{4}   & 49.6 & 72.4 & 74.5 & 83.3 & 72.4 & 29.1 & 70.2 & 64.5 \\
    \bottomrule
  \end{tabular}%
}
\vspace{-0.1in}
\end{table*}

\begin{table*}[htbp]
\centering
\caption{Evaluating base DeepSeek-Coder-V2-Lite-Instruct model Zero-Shot Results on downstream  Math Reasoning dataset, includes SingleEQ, MultiArith, AddSub, GSM8K, SVAMP, and AQuA.}
\label{tab:scale_laws_deepseek}
\vspace{-0.1in}
\resizebox{\textwidth}{!}{%
  \setlength{\tabcolsep}{1.2mm}% tighter columns for metrics
  \begin{tabular}{
    p{3.8cm}  % Model (wide, single multirow cell)
    p{3.0cm}  % Strategies (wider for readability)
    c         % Remain Experts (n')
    c         % Activate Experts (k')
    c c c c c c c c  % metrics
  }
    \toprule
    \textbf{Model} & \textbf{Strategies} & \textbf{Remain Experts ($n'$)} & \textbf{Activate Experts ($k'$)} &
    \textbf{GSM8k} & \textbf{SingleEq} & \textbf{SVAMP} & \textbf{MultiArith} &
    \textbf{AddSub} & \textbf{AQuA} & \textbf{mawps} & \textbf{AVG} \\
    \midrule

    %================== One big multirow for the Model column (17 rows total) ==================%
    \multirow{17}{*}{\textbf{DeepSeek-Coder-V2-Lite}} 
      & Base Model        & \textbf{64} & \textbf{6}  & 82.6 & 95.1 & 83.6 & 94.3 & 89.6 & 26.4 & 86.6 & 79.7 \\
    \cmidrule(l){2-12}
      & \multirow{6}{*}{Small Dense}
                           & \textbf{6}  & \textbf{6}  & 1.4  & 1.0  & 1.7  & 2.8  & 1.8  & 22.0 & 2.1  & 4.7 \\
      &                    & \textbf{12} & \textbf{12} & 1.7  & 11.0 & 4.5  & 7.2  & 13.2 & 24.0 & 9.7  & 10.2 \\
      &                    & \textbf{16} & \textbf{16} & 11.8 & 38.2 & 22.6 & 32.0 & 31.1 & 17.7 & 31.9 & 26.5 \\
      &                    & \textbf{20} & \textbf{20} & 26.0 & 63.2 & 46.0 & 67.7 & 56.5 & 17.7 & 58.4 & 47.9 \\
      &                    & \textbf{24} & \textbf{24} & 36.9 & 75.2 & 58.4 & 76.0 & 70.1 & 20.1 & 66.4 & 57.6 \\
      &                    & \textbf{32} & \textbf{32} & 47.8 & 83.5 & 71.4 & 88.2 & 81.8 & 24.4 & 79.6 & 68.1 \\
      &                    & \textbf{48} & \textbf{48} & 48.6 & 82.7 & 72.4 & 87.6 & 82.6 & 24.7 & 80.4 & 68.4 \\
    \cmidrule(l){2-12}
      & \multirow{5}{*}{Inference Reduce}
                           & \textbf{64} & \textbf{1}  & 28.5 & 68.7 & 50.7 & 73.0 & 64.6 & 24.0 & 72.6 & 54.6 \\
      &                    & \textbf{64} & \textbf{2}  & 49.2 & 86.6 & 70.5 & 92.8 & 79.2 & 24.8 & 78.2 & 68.8 \\
      &                    & \textbf{64} & \textbf{3}  & 54.3 & 87.2 & 76.6 & 92.3 & 80.3 & 25.6 & 84.9 & 71.6 \\
      &                    & \textbf{64} & \textbf{4}  & 53.7 & 87.0 & 76.1 & 94.7 & 83.5 & 26.4 & 83.2 & 72.1 \\
      &                    & \textbf{64} & \textbf{5}  & 57.9 & 89.0 & 79.8 & 95.7 & 84.6 & 25.2 & 84.5 & 73.8 \\
    \cmidrule(l){2-12}
      & \multirow{5}{*}{Small MoE}
                           & \textbf{12} & \textbf{6}  & 0.7  & 0.4  & 1.5  & 0.8  & 0.0  & 16.9 & 1.3  & 3.1 \\
      &                    & \textbf{16} & \textbf{6}  & 0.8  & 0.2  & 1.4  & 1.2  & 1.3  & 13.4 & 1.3  & 2.8 \\
      &                    & \textbf{24} & \textbf{6}  & 11.3 & 45.9 & 33.5 & 41.8 & 46.6 & 17.7 & 43.3 & 34.3 \\
      &                    & \textbf{32} & \textbf{6}  & 35.6 & 76.4 & 63.4 & 80.7 & 69.7 & 22.0 & 74.8 & 60.4 \\
      &                    & \textbf{48} & \textbf{6}  & 49.0 & 85.4 & 75.2 & 93.2 & 80.8 & 21.7 & 80.7 & 69.4 \\
    \bottomrule
  \end{tabular}%
}
\vspace{-0.1in}
\end{table*}

\clearpage
\newpage

\section{Selecting Top-$k$ Experts}
\label{appendix:experts_selection}
We define two top-\( k \) selection rules, selecting by magnitude score and selecting by activation ratio. Let $v^{(l)}_{i,j,k}$ be the output of expert $i$ at layer $l$ for token $k$ in sample $x_j$, with gate score $g^{(l)}_{i,j,k}$. Each sample has length $L_j$, and we draw a subset $\mathcal{D}_s=\{x_j\}_{j=1}^{N_s}$ from the training set. We compute a per-expert relevance score and pick Top-$k$ experts for routing or distillation.

\paragraph{Magnitude Score (ES-Mag).}
Estimate expert importance by average output magnitude:
\[
s^{(l)}_{i}=\frac{1}{N_s}\sum_{j=1}^{N_s}\frac{1}{L_j}\sum_{k=1}^{L_j}\left\|v^{(l)}_{i,j,k}\right\|.
\]

When only a scalar amplitude $\rho^{(l)}_{i,j,k}$ is available, we approximate $\|v^{(l)}_{i,j,k}\|\approx \rho^{(l)}_{i,j,k}$; this criterion favors experts with larger norm contributions (see Appendix~\ref{sec:theoretical_support_topk_select} for justification).

\paragraph{Activation Ratio (ES-Act).}
Estimate importance by how often the expert is selected:
\[
r^{(l)}_{i}=\frac{1}{N_s}\sum_{j=1}^{N_s}\frac{1}{L_j}\sum_{k=1}^{L_j}\frac{\mathbf{1}\!\left[g^{(l)}_{i,j,k}>0\right]}{K},
\]
where $K$ is the number of experts selected per token. This captures routing preference and data alignment.

\paragraph{Selection.}
For layer $l$, choose
\[
S^{(l)}=\mathrm{TopK}\big(\{s^{(l)}_{i}\}_{i}\big)\quad\text{or}\quad
S^{(l)}=\mathrm{TopK}\big(\{r^{(l)}_{i}\}_{i}\big),
\]
and select the variant by validation performance: \emph{ES-Mag} emphasizes magnitude-dominant contribution, while \emph{ES-Act} reflects gate-driven frequency. The ablation studies to investigate between Magnitude Score(ES-Mag) and Activation Ratio(ES-Act) can be found in Appendix~\ref{appendix:es-mag-versus-es-act}.

\clearpage
\newpage

\section{Auxiliary Loss for Load Balancing}
\label{appendix:aux_loss}

Uncontrolled routing strategies in Mixture-of-Experts (MoE) models often suffer from load imbalance. This manifests in two ways: (i) \emph{routing collapse}, where only a small subset of experts are consistently selected, leading to undertraining of the remaining experts; and (ii) \emph{computational imbalance}, where uneven routing across devices increases latency and reduces efficiency. To mitigate these issues, an auxiliary loss is commonly introduced in SOTA MoE models \citep{liu2024deepseek3, wang2024let}.

For a sequence of length $T$, the auxiliary loss is defined as
\[
\mathcal{L}_{\text{Balance}}
= \alpha \sum_{i=1}^{n} f_i P_i,
\]
where $\alpha$ is a hyperparameter controlling the strength of the regularization. Here,
\[
f_i = \frac{n}{KT} \sum_{t=1}^{T} \mathbf{1}\!\left[ g_{i,t} > 0 \right],
\qquad
P_i = \frac{1}{T}\sum_{t=1}^T s_{i,t}.
\]
The term $f_i$ measures the fraction of tokens routed to expert $i$, normalized by the total number of tokens $T$, experts $n$, and the per-token selection budget $K$. The term $P_i$ is the average routing probability assigned to expert $i$, where $s_{i,t}$ denotes the gating score of expert $i$ for token $t$. The loss encourages alignment between routing frequency ($f_i$) and gating probability ($P_i$), thereby preventing collapse and promoting balanced utilization of experts.

\clearpage
\newpage

\section{Auxiliary-Loss-Free Load Balancing Strategy}
\label{appendix:Auxiliary_Loss_Free}
To improve load balance without introducing an additional loss term, Deepseek-V3~\citep{liu2024deepseek3} adjust
the \emph{selection} rule by adding an expert-wise bias to the gating scores.
Let $s_{i,t}$ be the (un-normalized, before Softmax) gating score for expert $i$ on token $t$.
We define a biased score
\[
\tilde{s}_{i,t} \;=\; s_{i,t} + b_i,
\]
where $b_i$ is an expert-specific bias that is updated by a balancing controller
(e.g., based on utilization statistics; see remark below).  The Top-$k$
\emph{selection set} for token $t$ is then
\[
S_t \;=\; \mathrm{TopK}\bigl(\{\tilde{s}_{j,t}\}_{j=1}^n,\, K\bigr),
\qquad
g_{i,t} \;=\; \mathbf{1}[\,i \in S_t\,].
\]

\paragraph{Important distinction (selection vs.\ weighting).}
The bias $b_i$ is \emph{only} used to influence which experts enter $S_t$.
It does \emph{not} modify the routing weights used to combine expert outputs.
Weights are obtained from the \emph{unbiased} scores via softmax:
\[
w_{i,t} \;=\; \frac{\exp(s_{i,t})}{\sum_{j=1}^n \exp(s_{j,t})},
\]
and the token-level MoE output is
\[
y_t \;=\; \sum_{i=1}^n g_{i,t}\, w_{i,t}\, v_i
\;=\; \sum_{i \in S_t} w_{i,t}\, v_i,
\quad\text{with}\quad v_i=\rho_i e_i.
\]
Thus $b_i$ affects \emph{who is selected} but never changes the weights
$w_{i,t}$ applied to the selected experts in the forward pass.

\paragraph{Algorithm (per token $t$).}
\begin{enumerate}
    \item Compute unbiased scores $\{s_{i,t}\}_{i=1}^n$ and weights $w_{i,t}=\mathrm{softmax}(s_{i,t})$.
    \item Form biased scores $\tilde{s}_{i,t}=s_{i,t}+b_i$ and select $S_t=\mathrm{TopK}(\{\tilde{s}_{j,t}\}_{j=1}^n,K)$.
    \item Set indicators $g_{i,t}=\mathbf{1}[i\in S_t]$ and compute $y_t=\sum_{i\in S_t} w_{i,t} v_i$.
\end{enumerate}

\paragraph{Bias updating $b_i$.}
Any load-balancing controller can be used to update the biases; for example,
one may adjust $b_i$ as a function of the observed utilization $f_i$
and target utilization $K/n$ (e.g., with a moving-average estimator). In DeepSeek-V3, during training, they monitoring the expert load on the whole batch of each training step. At the end of each step, we will decrease the bias term by $\gamma$ if its corresponding expert is overloaded, and increase it by $\gamma$ if its corresponding expert is underloaded, where $\gamma$ is a hyper-parameter called bias update speed. Through the dynamic adjustment, DeepSeek-V3 keeps balanced expert load during
training, and achieves better performance than models that encourage load balance through
pure auxiliary losses.

\clearpage
\newpage

\section{Gradient Propagation Through Top-\(k\) Routing}
\label{appendix:esft_densemixer}
Consider a Mixture-of-Experts layer where the router produces gating scores $\{s_{i,t}\}_{i=1}^n$ for token $t$.  These scores are normalized via a softmax to obtain the routing weights
\[
w_{i,t} \;=\; \frac{\exp(s_{i,t})}{\sum_{j=1}^n \exp(s_{j,t})}\,,\qquad i=1,\dots,n.
\]
We then select the top-$k$ experts according to these weights and form the token-level output
\[
y_t \;=\; \sum_{i \in S_t} \, w_{i,t} \, v_i,
\]
where $S_t = \mathrm{TopK}\bigl(\{w_{j,t}\}_{j=1}^n\bigr)$ is the index set of the $k$ largest weights, and $v_i=\rho_i e_i$ is the contribution of expert $i$ (with $\rho_i$ the magnitude of $h_i$ and $e_i$ its normalized direction).  The final loss for this token is $\mathcal{L} = \mathcal{L}(y_t)$.

\paragraph{Gradient with respect to router parameters.}
Back-propagation through this layer requires differentiating the loss with respect to the router parameters $\theta$:
\[
\nabla_\theta \mathcal{L}
\;=\;
\sum_{i=1}^n
\left(\frac{\partial \mathcal{L}}{\partial y_t} \cdot v_i\right)
\cdot
\frac{\partial \,\mathrm{TopK}(w_{1,t},\dots,w_{n,t})_i}{\partial w_{i,t}}
\cdot
\frac{\partial w_{i,t}}{\partial \theta}.
\]
The Jacobian term $\partial w_{i,t}/\partial\theta$ is determined by the softmax of the gating scores $s_{i,t}$, while the middle factor contains the (non-differentiable) Top-$k$ selection.

\paragraph{Conventional approximation: SFT and ESFT}
A common approximation in MoE training treats the Top-$k$ operation as if it were differentiable by passing gradients only through the selected experts.  Formally, one replaces
\[
\frac{\partial\,\mathrm{TopK}(w_{1,t},\dots,w_{n,t})_i}{\partial w_{j,t}}
\;\approx\;
\delta_{i,j}\,\mathbf{1}[i\in S_t],
\]
where $\delta_{i,j}$ is the Kronecker delta.  Under this approximation, the router gradient reduces to
\[
\nabla_\theta \mathcal{L}
\;\approx\;
\sum_{i\in S_t}
\left(\frac{\partial \mathcal{L}}{\partial y_t}\cdot v_i\right)
\,\frac{\partial w_{i,t}}{\partial \theta}.
\]
Thus only the experts chosen in $S_t$ receive gradient updates through the gating mechanism.

\paragraph{Straight-through (STE) approximation: DenseMixer}
An alternative, more precise approximation—used in methods like DenseMixer—employs a straight-through estimator (STE).  In the backward pass, the Top-$k$ operation is treated as the identity map:
\[
\frac{\partial\,\mathrm{TopK}(w_{1,t},\dots,w_{n,t})_i}{\partial w_{j,t}}
\;\approx\;\delta_{i,j}.
\]
This allows gradients to flow to all experts’ routing weights, yielding
\[
\nabla_\theta \mathcal{L}
\;\approx\;
\sum_{i=1}^n
\left(\frac{\partial \mathcal{L}}{\partial y_t}\cdot v_i\right)
\,\frac{\partial w_{i,t}}{\partial \theta}.
\]
In this view, the forward pass still uses a hard Top-$k$ selection, but the backward pass distributes gradients as though the selection were an identity operator.

\medskip
\noindent\textbf{Summary.}  The conventional method restricts gradient updates to the selected experts $S_t$, while the straight-through method propagates gradients to all experts by overriding the Top-$k$ operation in the backward pass.

\clearpage
\newpage

\section{Theoretical Supports for Top-K Selection}
\label{sec:theoretical_support_topk_select}
In a Mixture-of-Experts (MoE) architecture, each expert contributes to the overall output as
\[
v_i = \rho_i e_i,
\]
where \(\rho_i\) is the gating weight corresponds to the \( i \)-th expert, \(e_i\) is the expert output, and $v_i$ is the output vector of the $i$-th expert. In a dense model, the final output is given by

\[
y = \sum_{i=1}^n v_i = \sum_{i=1}^n \rho_i e_i.
\]
In a sparse Mixture-of-Experts (MoE) model, we aim to reduce computation by selecting only a subset of experts. Thus, we wish to approximate \( y \) using
\[
\hat{y} = \sum_{i \in S} v_i,
\]
where \( S \) is a subset of indices with \( |S| = k \ll n \). This objective can be formulated as the minimization problem
\[
\min_{\lambda_1, \ldots, \lambda_n \in \{0,1\}} \left\|\sum_{i=1}^n v_i - \sum_{i=1}^n \lambda_i\, v_i \right\|^2 \quad \text{subject to} \quad \sum_{i=1}^n \lambda_i = k,
\]
where \( \lambda_i = 1 \) indicates that expert \( i \) is selected, and \( \lambda_i = 0 \) indicates it is omitted. In many practical scenarios, especially when the normalized directions \( e_i \) are not strongly correlated, this minimization is well approximated by selecting the experts with the largest values of \( \rho_i \). Intuitively, experts with large \( \rho_i \) contribute most significantly to the norm of \( y \), so preserving these in the approximation yields a smaller error. We analyze why selecting experts with large \( \rho_i \) is a reasonable approximation in the following.

In a Mixture-of-Experts (MoE) architecture, each expert contributes to the overall output as
\[
v_i = \rho_i e_i,
\]
where \(\rho_i\) is the gating weight and \(e_i\) is the expert output. When analyzing why the top-\( k \) selection rule arises, it is instructive to consider two scenarios: one in which the vectors \( v_i \) are orthonormal (or nearly so) and another in which they have general correlations.

\noindent In this appendix, we show that in the non-orthonormal case, selecting the top-\(k\) experts with the largest \(\rho_i\) provides a close approximation to the full model output while substantially reducing computational cost. In the orthonormal case, this selection is provably optimal; in the general case, it serves as a widely used and effective heuristic.

\subsection*{The Orthonormal (or Weakly-Correlated) Case}

Assume that the vectors \( v_1, v_2, \dots, v_n \) are strictly orthonormal, i.e.,
\[
v_i^\top v_j =
\begin{cases}
0, & \text{if } i \neq j, \\
\|v_i\|^2, & \text{if } i = j.
\end{cases}
\]
Then, the squared norm of the omitted portion,
\[
\left\|\sum_{i=1}^n (1-\lambda_i)v_i \right\|^2,
\]
expands as
\[
\left\|\sum_{i=1}^n (1-\lambda_i)v_i \right\|^2 = \sum_{i=1}^n (1-\lambda_i)^2 \|v_i\|^2,
\]
and since \(\lambda_i \in \{0,1\}\), we have \((1-\lambda_i)^2 = (1-\lambda_i)\). Therefore, the objective becomes
\[
\sum_{i=1}^n (1-\lambda_i) \|v_i\|^2,
\]
subject to \(\sum_{i=1}^n \lambda_i = k\). To minimize this quantity, it is optimal to set \(\lambda_i = 1\) for the \( k \) vectors with the largest norms \(\|v_i\|^2\) and \(\lambda_i = 0\) for the others. In the orthonormal case, this strategy is provably optimal.

\medskip

\noindent Even if the vectors are only weakly correlated, the same principle generally holds: larger magnitudes imply a larger contribution to the overall sum, so omitting vectors with small \(\|v_i\|\) results in a minor error, making the top-\( k \) selection by magnitude a robust heuristic.

\subsection*{The General (Non-Orthonormal) Case}

When the vectors \( v_i \) have significant correlations, the cross terms do not vanish. In this case, the error term becomes
\[
\left\|\sum_{i=1}^n (1-\lambda_i)v_i \right\|^2 = \sum_{i=1}^n (1-\lambda_i) \|v_i\|^2 + 2\sum_{1 \leq i < j \leq n} (1-\lambda_i)(1-\lambda_j) v_i^\top v_j.
\]
Here, the cross terms \( v_i^\top v_j \) can affect the error significantly. In principle, finding the subset \( S \) that minimizes this expression exactly is an NP-hard combinatorial problem. However, in practice, one commonly uses the heuristic of selecting the top \( k \) experts based on the individual magnitudes \( \|v_i\| \) (or a predicted magnitude \(\rho_i\)). This approach is effective because, in many settings, the largest magnitude vectors still dominate the overall contribution even when correlations are present. In scenarios where two high-magnitude vectors are strongly correlated, more sophisticated selection methods might improve the approximation, but the top-\( k \) rule remains a strong and computationally efficient baseline.

\medskip

\noindent \textbf{Conclusion:} Whether the expert output vectors are orthonormal or generally correlated, the top-\( k \) selection rule emerges from the objective of preserving the dominant contributions to the sum while minimizing approximation error. In an MoE architecture, each expert's output \( v_i = \rho_i e_i \) contributes to the overall sum. By selecting the \( k \) experts with the largest \( \rho_i \), one can achieve a good approximation of the full model output with significantly reduced computational cost. In the orthonormal case, this method is exactly optimal, while in the general case it remains a widely-used and effective heuristic.

\clearpage
\newpage

\section{Gradient Lower Bound for Condenser Experts}
\label{sec:proof_persistent_grad}

\begin{proof}
We consider a Mixture-of-Experts model of the form
\[
h(x)
=
\sum_{i \in S(x)}
g_i(x)\,E_i(x;\theta_i)
+
\sum_{k=1}^{\hat{n}} E_k(x),
\]
with expected risk
\[
\mathcal{L}(\theta)
=
\mathbb{E}_{(x,y)\sim\mathcal{D}}
\big[
\ell(h(x),y)
\big].
\]

Let
\[
\delta(x) := \frac{\partial \ell}{\partial h(x)}
\]
denote the backpropagated gradient at the MoE output.

\paragraph{Gradient identity.}
Since $j \in J \subseteq S(x)$ for all $x$, the model output depends on $\theta_j$ through the term
\[
g_j(x)\,E_j(x;\theta_j).
\]
By the chain rule,
\[
\nabla_{\theta_j} h(x)
=
g_j(x)\,
\frac{\partial E_j(x;\theta_j)}{\partial \theta_j}.
\]
Therefore,
\[
\nabla_{\theta_j}\mathcal{L}
=
\mathbb{E}_{(x,y)\sim\mathcal{D}}
\Big[
\delta(x)\,
g_j(x)\,
\frac{\partial E_j(x;\theta_j)}{\partial \theta_j}
\Big].
\]

\paragraph{Lower bound.}
We now derive the gradient magnitude lower bound.
Consider the event
\(
\{g_j(x)\ge \epsilon\}.
\)
Then
\[
\mathbb{E}
\big[
\|\nabla_{\theta_j}\mathcal{L}\|
\big]
=
\mathbb{E}
\Big[
\big\|
\delta(x)\,
g_j(x)\,
\frac{\partial E_j(x)}{\partial \theta_j}
\big\|
\Big].
\]

Restricting to the event $\{g_j(x)\ge \epsilon\}$ gives
\[
\ge
\mathbb{E}
\Big[
g_j(x)\,
\big\|
\delta(x)\,
\frac{\partial E_j(x)}{\partial \theta_j}
\big\|
\mathbf{1}_{\{g_j(x)\ge\epsilon\}}
\Big].
\]

Using $g_j(x)\ge\epsilon$ on this event,
\[
\ge
\epsilon\,
\mathbb{P}\big(g_j(x)\ge\epsilon\big)\,
\mathbb{E}
\Big[
\big\|
\delta(x)\,
\frac{\partial E_j(x)}{\partial \theta_j}
\big\|
\;\big|\;
g_j(x)\ge\epsilon
\Big].
\]

By assumption,
\(
\mathbb{P}(g_j(x)\ge\epsilon)\ge p,
\)
which yields
\[
\mathbb{E}
\big[
\|\nabla_{\theta_j}\mathcal{L}\|
\big]
\ge
p\,\epsilon\,
\mathbb{E}
\Big[
\big\|
\delta(x)\,
\frac{\partial E_j(x)}{\partial \theta_j}
\big\|
\;\big|\;
g_j(x)\ge\epsilon
\Big].
\]
\end{proof}

% \section{Definition of Correlation}
% \label{appendix:correlation_define}
% \paragraph{Correlation Gain on \texttt{down\_proj} (vs.\ Base).}

% For each expert $e$, let
% $\mathbf{w}_e=\mathrm{vec}(W^{(e)}_{\text{down}})$ and define
% \[
% \mathcal{C}_{ij}
% =\mathrm{Corr}(\mathbf{w}_i,\mathbf{w}_j)
% =\frac{(\mathbf{w}_i-\bar{w}_i\mathbf{1})^\top(\mathbf{w}_j-\bar{w}_j\mathbf{1})}
% {\|\mathbf{w}_i-\bar{w}_i\mathbf{1}\|_2\;\|\mathbf{w}_j-\bar{w}_j\mathbf{1}\|_2},
% \]
% where $\bar{w}_i$ is the mean of $\mathbf{w}_i$.  
% Across three settings $s\!\in\!\{1,2,3\}$ (\,$1$: Base,\; $2$: ESFT,\; $3$: Expert~Condenser\,),
% we compute the pairwise percent gain over Base for \emph{shared$\to$regular} pairs:
% \[
% \Delta^{(l)}_{ij,s}(\%) \;=\;
% 100\times\frac{\mathcal{C}^{(l)}_{ij,s}-\mathcal{C}^{(l)}_{ij,1}}{\mathcal{C}^{(l)}_{ij,1}},
% \quad i\in\{0,1\},\; j\in\{2,\dots,63\},\; s\in\{2,3\},
% \]
% where $l$ indexes the MoE layer. A layer-level summary is
% \[
% \bar{\Delta}^{(l)}_{s}(\%) \;=\;
% \frac{1}{2\cdot 62}\sum_{i=1}^{2}\sum_{j=3}^{64}\Delta^{(l)}_{ij,s}(\%),
% \qquad s\in\{2,3\},
% \]
% and the aggregate across all layers $L$ is
% \[
% \bar{\Delta}_{s}(\%) \;=\; \frac{1}{L}\sum_{l=1}^{L}\bar{\Delta}^{(l)}_{s}(\%),
% \qquad
% \sigma_{s} \;=\; \mathrm{Std}\!\left(\{\bar{\Delta}^{(l)}_{s}(\%)\}_{l=1}^{L}\right).
% \]

\clearpage
\newpage

\section{Out-of-Domain Generalization}
\label{appendix:ood_generalization}

To further evaluate the generalization ability of ExpertCondenser beyond the target training domain, we conduct an additional out-of-domain experiment. 
In this setting, we fine-tune the OLMoE-7B model on a \textbf{commonsense reasoning dataset}, and evaluate the resulting models on \textbf{mathematical reasoning benchmarks}, including GSM8K, SingleEq, SVAMP, MultiArith, AddSub, AQuA, and MAWPS.

We note that, similar to prior work such as DenseMixer~\cite{yao2025densemixer}, our method is primarily designed to improve post-training performance on a given target task, rather than to preserve broad multi-domain capabilities during supervised fine-tuning. 
Nevertheless, this experiment provides additional insight into how ExpertCondenser behaves under cross-domain evaluation.

Table~\ref{tab:ood_generalization} reports the results. 
ExpertCondenser achieves performance comparable to SFT and ESFT across all benchmarks, and consistently outperforms DenseMixer. 
While all fine-tuning methods exhibit degradation relative to the base model when evaluated on an unrelated domain, ExpertCondenser does not introduce additional degradation compared to standard SFT-based approaches. These results indicate that ExpertCondenser maintains similar cross-domain generalization behavior to existing post-training methods. 
Despite enforcing structured sparsity and persistent expert activation, the model retains sufficient flexibility to support out-of-domain evaluation without additional performance loss.

\begin{table*}[h]
\centering
\small
\resizebox{\textwidth}{!}{%
\begin{tabular}{l c c c c c c c c c c}
\toprule
\textbf{FT Dataset} & \textbf{Model} & \textbf{Post-train Type} 
& \textbf{GSM8K} & \textbf{SingleEq} & \textbf{SVAMP} 
& \textbf{MultiArith} & \textbf{AddSub} & \textbf{AQuA} & \textbf{MAWPS} & \textbf{AVG} \\
\midrule
\multirow{5}{*}{Commonsense} 
& \multirow{5}{*}{OLMoE-7B} 
& Base Model & 16.1 & 23.6 & 17.7 & 9.2 & 21.3 & 22.8 & 13.9 & \textbf{17.8} \\
& & ExpertCondenser (Ours) & 8.9 & 13.7 & 11.6 & 5.4 & 13.8 & 18.6 & 11.3 & 11.9 \\
& & SFT & 6.8 & 14.6 & 12.5 & 5.7 & 13.4 & 17.8 & 8.6 & 11.3 \\
& & ESFT & 11.2 & 15.6 & 13.7 & 4.7 & 12.4 & 16.7 & 7.9 & 11.7 \\
& & DenseMixer & 2.3 & 5.9 & 6.1 & 3.7 & 6.1 & 17.6 & 2.9 & 6.4 \\
\bottomrule
\end{tabular}}
\caption{Out-of-domain evaluation: models fine-tuned on commonsense reasoning and evaluated on mathematical reasoning tasks. 
ExpertCondenser performs comparably to SFT and ESFT, and consistently outperforms DenseMixer.}
\label{tab:ood_generalization}
\end{table*}

\paragraph{Generalization after tuning on general instruction data.}
The experiment above tunes on one reasoning domain and evaluates on another. A stricter probe removes reasoning supervision from the training data entirely and asks how much of the base model's math ability survives fine-tuning. We fine-tune Qwen1.5-MoE-A2.7B on 10K general-instruction examples from \texttt{tatsu-lab/alpaca}, which contain no mathematical data, and evaluate on the same seven math benchmarks under a matched two-epoch budget for both methods.
Table~\ref{tab:alpaca_ood} reports the results. ExpertCondenser retains a $1.60$-point higher average than SFT ($33.06$ vs.\ $31.46$). The advantage is not uniform---SFT leads on GSM8K, MultiArith, and AQuA---but ExpertCondenser leads on the remaining four benchmarks and on the average. This is consistent with the router-KL measurement in \S~\ref{subsection:routing_and_output_analysis}: ExpertCondenser departs least from the base model's routing distribution, so fine-tuning on an unrelated domain overwrites less of the base model's behavior.

\begin{table*}[h]
\centering

\small
\adjustbox{max width=\textwidth}{%
\begin{tabular}{l c c c c c c c c c c}
\toprule
\textbf{FT Dataset} & \textbf{Model} & \textbf{Post-train Type}
& \textbf{GSM8K} & \textbf{SingleEq} & \textbf{SVAMP}
& \textbf{MultiArith} & \textbf{AddSub} & \textbf{AQuA} & \textbf{MAWPS} & \textbf{AVG} \\
\midrule
\multirow{2}{*}{Alpaca-10K}
& \multirow{2}{*}{Qwen1.5-MoE-A2.7B}
& SFT & \textbf{14.4} & 49.6 & 33.2 & \textbf{13.2} & 38.0 & \textbf{26.8} & 45.0 & 31.46 \\
& & ExpertCondenser (Ours) & 11.7 & \textbf{55.9} & \textbf{37.8} & 9.5 & \textbf{46.3} & 24.0 & \textbf{46.2} & \textbf{33.06} \\
\bottomrule
\end{tabular}}
\caption{Out-of-domain evaluation after fine-tuning on general instruction data: Qwen1.5-MoE-A2.7B tuned on 10K \texttt{tatsu-lab/alpaca} examples (no mathematical supervision) and evaluated on the seven math benchmarks, with a matched two-epoch budget.}
\label{tab:alpaca_ood}
\end{table*}

\clearpage
\newpage

\section{Expert Selection and Bias Design}
\label{appendix:bias_selection}

Our routing design follows the auxiliary-loss-free load-balancing principle introduced in DeepSeek~\cite{wang2024auxiliary}, where the bias term is applied only to routing decisions, and not to the computation of gating weights. This separation is critical for both stability and performance. 
Given routing logits $s_i$, we introduce a bias term $b_i$ and perform Top-$k$ selection using
\[
\mathrm{TopK}(\{s_i + b_i\}),
\]
while the gating weights $g_{i,t}$ are computed from the original logits $s_i$ without bias. 
As stated in DeepSeek~\cite{wang2024auxiliary}, the bias term is used solely to adjust routing behavior and is not included in the computation of expert output weights.

\paragraph{(1) Bias controls selection, not representation.}
The bias term $b_i$ is designed to influence \emph{which experts are selected}, rather than \emph{how much each expert contributes}. 
By applying bias only before Top-$k$ selection, we modify the routing pattern while preserving the original gating function for combining expert outputs.

If the bias were also added to the gating weights (e.g., inside the softmax or sigmoid), it would directly distort the magnitude of expert contributions. 
This would alter the representation distribution across layers, introduce additional gradients into expert outputs, and interfere with the learned semantics of the routing function.

\paragraph{(2) Preserving stable gating geometry.}
The router logits are typically computed as
\[
s_i = G(x^\top e_i),
\]
where $e_i$ denotes expert embeddings. 
Injecting bias into the gating weights would modify the geometry of this scoring function, shifting expert centroids and distorting the distribution of routing probabilities.

DeepSeek~\cite{wang2024auxiliary} explicitly avoids this design, as modifying gating weights introduces undesired gradients and destabilizes training. 
Their formulation emphasizes that auxiliary-free balancing should control load without interfering with the learned gating distribution.

\paragraph{(3) Avoiding gradient interference.}
If the bias term were included in the gating weights, it would affect both selection and weighting simultaneously. 
In this case, bias updates would propagate through expert outputs, mixing routing control with model optimization. 
This coupling introduces additional gradient interference, which can destabilize training and degrade performance.

In contrast, our design isolates routing control (via $b_i$) from representation learning (via $g_{i,t}$), maintaining a clean separation between selection and computation.

DeepSeek~\cite{wang2024auxiliary} reports that incorporating bias directly into gating weights leads to unstable gate distributions and worse performance (see Table 4 in their paper). 
We adopt the same design principle in ExpertCondenser.

In addition, we observe that using $s_i + b_i$ for selection produces a substantially different expert ranking compared to using $s_i$ alone, particularly in later layers where bias accumulation drives structured sparsity. 
This difference reflects the intended effect of bias-based routing: reshaping expert utilization without perturbing the learned gating function.

\clearpage
\newpage

\section{Expert Correlation Analysis}
\label{appendix:expert_correlation_analysis}

To assess how fine-tuning alters dependencies between experts, we examine the similarity of their parameter updates. Specifically, we compute the Pearson correlation between the parameters of the condensed experts and those of other experts, comparing the fine-tuned model to the base model. This measure captures linear relationships in parameter changes, allowing us to track how post-training reshapes inter-expert dependencies.
A formal definition of Pearson correlation is provided in Appendix~\ref{appendix:correlation_define}.

\begin{figure}[h]
  \centering
  \begin{subfigure}[t]{0.48\linewidth}
    \centering
    \includegraphics[width=\linewidth]{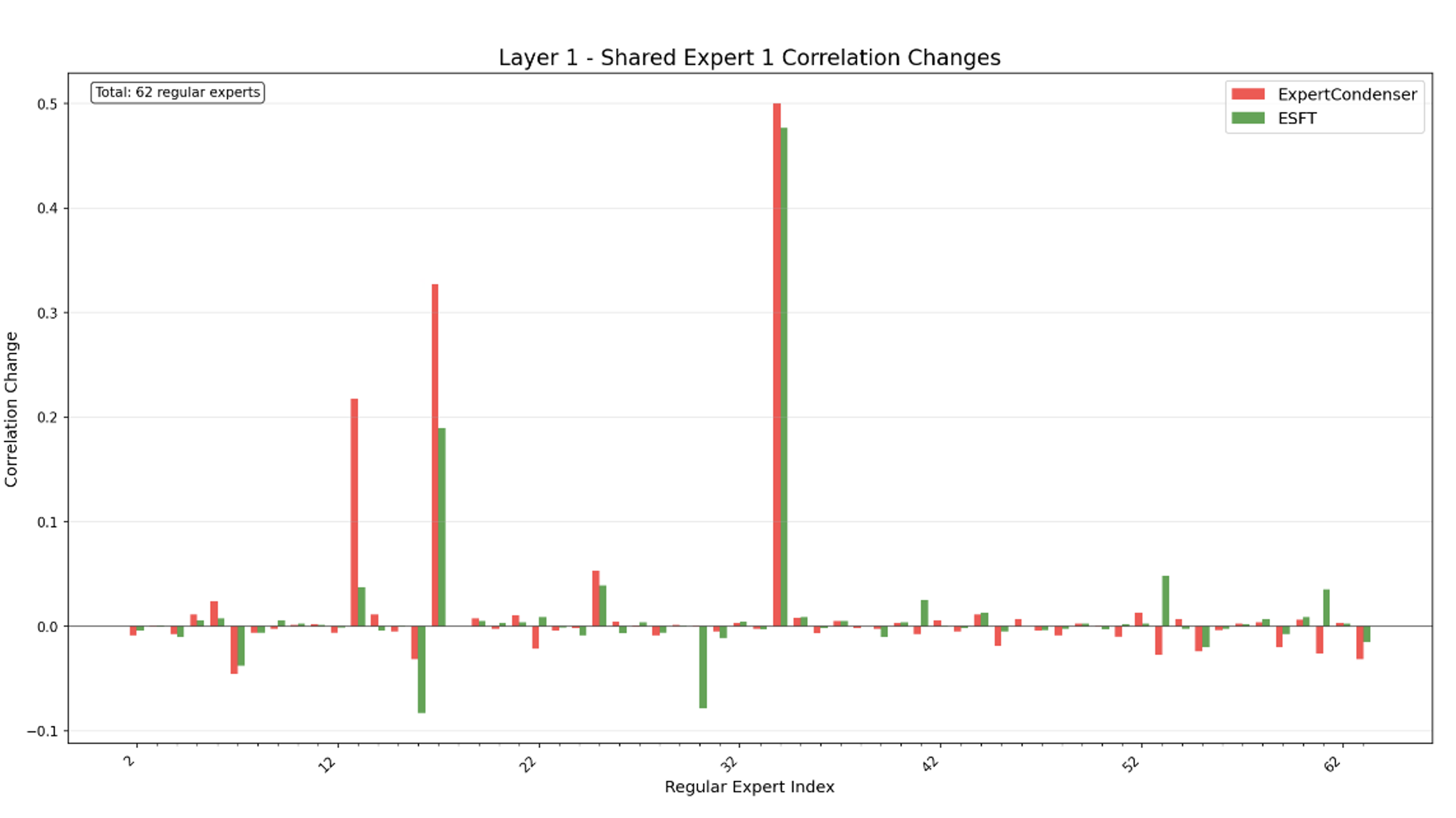}
    \caption{Shared Expert 0 correlation changes.}
    \label{fig:shared0}
  \end{subfigure}\hfill
  \begin{subfigure}[t]{0.48\linewidth}
    \centering
    \includegraphics[width=\linewidth]{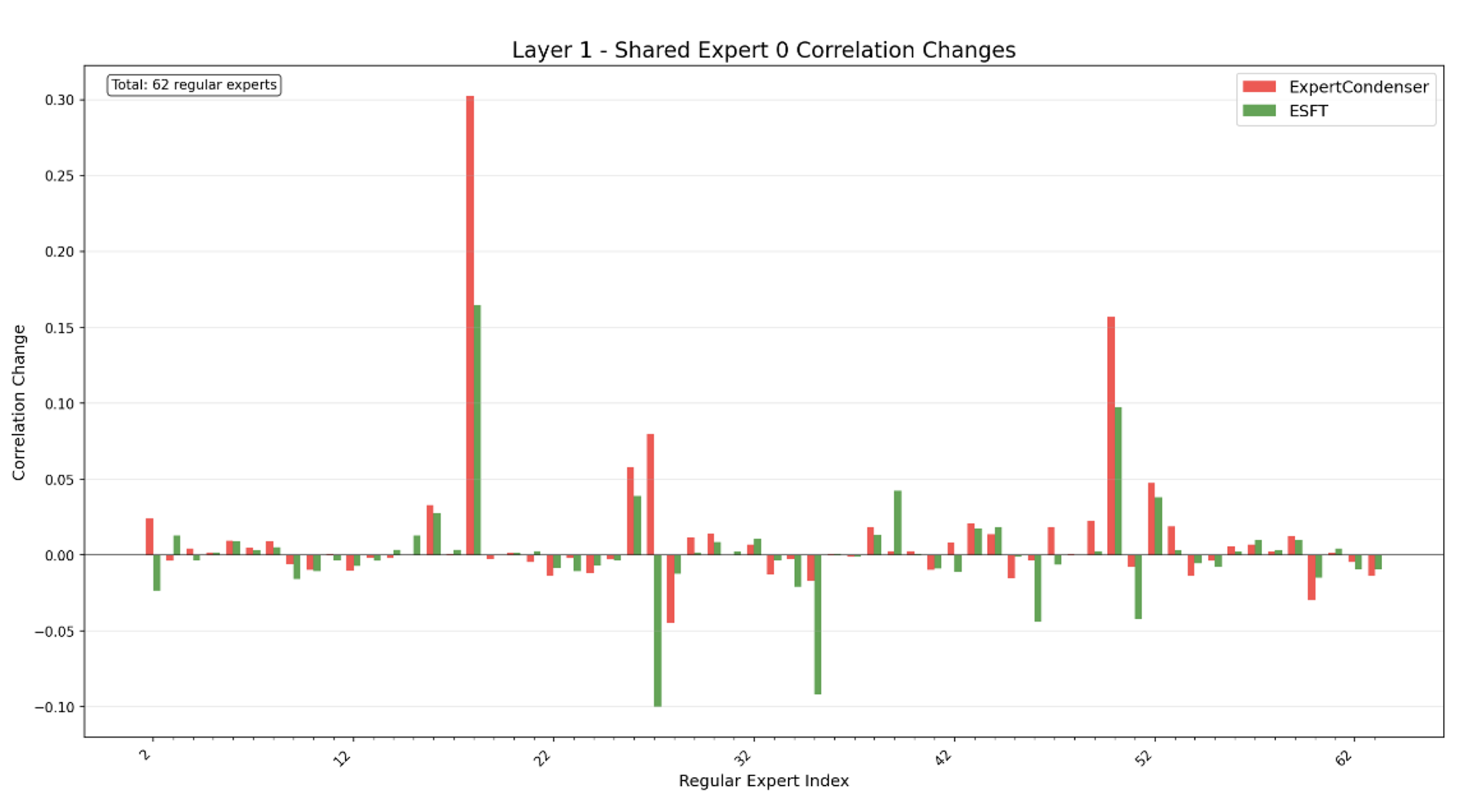}
    \caption{Shared Expert 1 correlation changes.}
    \label{fig:shared1}
  \end{subfigure}
  \caption{Correlation changes between the shared expert and regular experts at Layer 1. Each bar shows how the correlation between a regular expert and the shared expert changes after fine-tuning, compared to the base model. ExpertCondenser (red) causes concentrated shifts, markedly strengthening positive correlations while suppressing negative ones. In contrast, ESFT (green) produces smaller, more diffuse adjustments across experts. These results illustrate that ExpertCondenser induces larger and more concentrated correlation shifts, while ESFT exerts a milder influence.}
  \label{fig:shared-corr}
\end{figure}

\textbf{Observation.} Condenser experts consolidate the information, resulting in pronounced, concentrated shifts. Regularly routed experts are compared against these shared experts by computing Pearson correlations between their down-projection weight matrices before and after fine-tuning. Relative changes in these correlations—normalized by the base model—reflect how each method reshapes inter-expert dependencies. As shown in Fig. \ref{fig:shared-corr}, both ExpertCondenser and ESFT increase correlations relative to the baseline, with ExpertCondenser driving a stronger global increase in correlations. Relative correlation increases average 0.005 more across all layers compared to ESFT. The earliest layer shows the most significant shifts, highlighting ExpertCondenser’s central role in shaping foundational representations.

\subsection{Definition of Correlation}
\label{appendix:correlation_define}

\paragraph{Correlation on \texttt{down\_proj} Weights.}

Consider an MoE layer with a set of experts
\[
\mathcal{E} = \{1,\dots,E\},
\]
partitioned into a \emph{shared expert set} $\mathcal{E}_{\mathrm{sh}}$ and a
\emph{regular expert set} $\mathcal{E}_{\mathrm{reg}}$, with
$\mathcal{E}_{\mathrm{sh}}\cap\mathcal{E}_{\mathrm{reg}}=\emptyset$,
$\mathcal{E}_{\mathrm{sh}}\cup\mathcal{E}_{\mathrm{reg}}=\mathcal{E}$.

For expert $e\!\in\!\mathcal{E}$ under setting $s$, let
\[
\mathbf{w}^{(l)}_{e,s}
=
\mathrm{vec}\!\left(W^{(l,e)}_{\mathrm{down},s}\right)
\in \mathbb{R}^d,
\]
and denote its sample mean by
$
\bar{w}^{(l)}_{e,s}
= \frac{1}{d}\sum_{k=1}^d (\mathbf{w}^{(l)}_{e,s})_k.
$
Define the centered weight vector
\[
\tilde{\mathbf{w}}^{(l)}_{e,s}
=
\mathbf{w}^{(l)}_{e,s}
-
\bar{w}^{(l)}_{e,s}\mathbf{1}.
\]

The Pearson correlation between experts $i$ and $j$ in layer $l$ under setting
$s$ is then
\[
\mathcal{C}^{(l)}_{ij,s}
=
\frac{
\left\langle
\tilde{\mathbf{w}}^{(l)}_{i,s},
\tilde{\mathbf{w}}^{(l)}_{j,s}
\right\rangle
}{
\left\|
\tilde{\mathbf{w}}^{(l)}_{i,s}
\right\|_{2}
\left\|
\tilde{\mathbf{w}}^{(l)}_{j,s}
\right\|_{2}
}.
\]

We study three settings:
\[
s = 1:\text{Base}, \qquad
s = 2:\text{ESFT}, \qquad
s = 3:\text{ExpertCondenser}.
\]

\paragraph{Percent Correlation Gain.}
For each shared expert $i\!\in\!\mathcal{E}_{\mathrm{sh}}$ and each regular expert
$j\!\in\!\mathcal{E}_{\mathrm{reg}}$, the percent change in correlation relative
to Base is defined as
\[
\Delta^{(l)}_{ij,s}(\%)
=
100\,
\frac{
\mathcal{C}^{(l)}_{ij,s}
-
\mathcal{C}^{(l)}_{ij,1}
}{
\left|\mathcal{C}^{(l)}_{ij,1}\right|
},
\qquad
s \in \{2,3\}.
\]
This “relative effect size” formulation stabilizes interpretation when the Base
correlation is negative or near zero.

\paragraph{Layer-Level Aggregation.}
The average correlation gain for setting $s$ in layer $l$ is
\[
\bar{\Delta}^{(l)}_{s}(\%)
=
\frac{1}{
\lvert\mathcal{E}_{\mathrm{sh}}\rvert \cdot
\lvert\mathcal{E}_{\mathrm{reg}}\rvert
}
\sum_{i\in\mathcal{E}_{\mathrm{sh}}}
\sum_{j\in\mathcal{E}_{\mathrm{reg}}}
\Delta^{(l)}_{ij,s}(\%).
\]

\paragraph{Model-Level Summary Across $L$ Layers.}
Define the overall average and variability as
\[
\bar{\Delta}_{s}(\%)
=
\frac{1}{L}
\sum_{l=1}^{L}
\bar{\Delta}^{(l)}_{s}(\%),
\qquad
\sigma_{s}
=
\mathrm{Std}
\!\left(
\left\{
\bar{\Delta}^{(l)}_{s}(\%)
\right\}_{l=1}^{L}
\right).
\]
These summarize how strongly fine-tuning reshapes global correlations between
shared experts and regular experts across the entire model.

\clearpage
\newpage

\section{Math7K Dataset}
\label{appendix:math7k}

\texttt{Math10K} dataset can evaluate the effectiveness of LLMs on the arithmetic reasoning task. \texttt{Math10K} incorporate six subsets including \texttt{GSM8k}, \texttt{SingleEq}, \texttt{SVAMP}, \texttt{MultiArith}, \texttt{AddSub}, and \texttt{AQuA}.(1) the \texttt{GSM8K}~\citep{cobbe2021gsm8k} dataset consists of high quality linguistically diverse grade school math word problems created by human problem
writers, (2) the \texttt{SVAMP}~\citep{patel-etal-2021-nlp} benchmark consists of one-unknown arithmetic word
problems for up-to-4 grade level students by making simple changes to a set of problems from another existing dataset, (3) the \texttt{MultiArith}~\citep{roy2016solving} dataset of math word problems requiring multiple reasoning steps and operations, (4) the \texttt{AddSub}~\citep{hosseini2014learning} dataset of addition and subtraction arithmetic word problems,
(5) the \texttt{AQuA}~\citep{ling2017program} dataset of algebraic word problems with natural language rationales, and (6) the \texttt{SingleEq}~\citep{koncel2015parsing} dataset of grade-school algebra word problems that map to single equations with varying length;

% \section{MoE Expert Sparsity after Post-training}

% In this appendix, we compare the sparsity of activated expert after different post-training methods.

% \section{Bias overlap with token activated frequency}

% We study the overlap between positive bias and the frequency of expert being activated during inference.

\section{ES-Act versus ES-Mag}
\label{appendix:es-mag-versus-es-act}
In this Appendix, we conduct ablation studies to investigate between Magnitude Score(ES-Mag) and Activation Ratio(ES-Act), which one is the better metric to select preserving experts when we converting the original Mixture of Expert model into smaller models (either smaller dense models or smaller MoE models).

\begin{table*}[htbp]
\centering
\caption{Evaluation of post-trained models (Zero-Shot results) on downstream Math Reasoning datasets, including SingleEQ, MultiArith, AddSub, GSM8K, SVAMP, and AQuA.}
% Renamed: this label previously duplicated the main-text ablation table in Sec. 5.4,
% which made every \ref{tab:mix_dataset_results} resolve to this appendix table.
\label{tab:zero_shot_math_appendix}
\vspace{-0.1in}

\resizebox{\textwidth}{!}{%
\setlength{\tabcolsep}{1.5mm}%
\begin{tabular}{p{1.5cm} p{2.5cm} c c c c c c c c c}
\toprule
\textbf{Metric} 
& \textbf{Method} 
& \textbf{Remaining Experts} 
& \textbf{GSM8k}
& \textbf{SingleEq}
& \textbf{SVAMP} 
& \textbf{MultiArith} 
& \textbf{AddSub} 
& \textbf{AQuA} 
& \textbf{mawps} 
& \textbf{AVG} \\
\midrule

%===== ES-Act =====
\multirow{11}{*}{\textbf{ES-Act}} 

& \multirow{6}{*}{\textbf{Smaller-Dense}} 
& 6
& 1.4 & 1.0 & 1.7 & 2.8 & 1.8 & 22.0 & 2.1 & 4.7 \\

& & 12
& 1.7 & 11.0 & 4.5 & 7.2 & 13.2 & 24.0 & 9.7 & 10.2 \\

& & 16
& 11.8 & 38.2 & 22.6 & 32.0 & 31.1 & 17.7 & 31.9 & 26.5 \\

& & 20
& 26.0 & 63.2 & 46.0 & 67.7 & 56.5 & 17.7 & 58.4 & 48.0 \\

& & 24
& 36.9 & 75.2 & 58.4 & 76.0 & 70.1 & 20.1 & 66.4 & 57.6 \\

& & 32
& 47.8 & 83.5 & 71.4 & 88.2 & 81.8 & 24.4 & 79.6 & 68.1 \\

\cmidrule(l){2-11}

& \multirow{5}{*}{\textbf{Smaller-MoE}} 

& 12
& 0.7 & 0.4 & 1.5 & 0.8 & 0.0 & 16.9 & 1.3 & 3.1 \\

& & 16
& 0.8 & 0.2 & 1.4 & 1.2 & 1.3 & 13.4 & 1.3 & 2.8 \\

& & 24
& 11.3 & 45.9 & 33.5 & 41.8 & 46.6 & 17.7 & 43.3 & 34.3 \\

& & 32
& 35.6 & 76.4 & 63.4 & 80.7 & 69.7 & 22.0 & 74.8 & 60.4 \\

& & 48
& 49.0 & 85.4 & 75.2 & 93.2 & 80.8 & 21.7 & 80.7 & \textbf{69.4} \\

\midrule

%===== ES-Mag =====
\multirow{11}{*}{\textbf{ES-Mag}} 

& \multirow{6}{*}{\textbf{Smaller-Dense}} 

& 6
& 1.6 & 1.2 & 2.1 & 2.6 & 2.1 & 18.9 & 2.3 & 4.4 \\

& & 12
& 1.8 & 11.8 & 5.2 & 6.8 & 13.6 & 23.8 & 9.4 & 10.3 \\

& & 16
& 12.6 & 38.8 & 23.8 & 33.8 & 32.6 & 22.4 & 31.9 & 28.0 \\

& & 20
& 25.4 & 64.8 & 45.6 & 68.8 & 56.3 & 18.2 & 59.8 & 48.4 \\

& & 24
& 37.4 & 76.8 & 60.2 & 75.4 & 70.8 & 18.8 & 65.2 & 57.8 \\

& & 32
& 48.2 & 82.8 & 72.2 & 87.8 & 82.2 & 24.2 & 83.2 & 68.6 \\

\cmidrule(l){2-11}

& \multirow{5}{*}{\textbf{Smaller-MoE}} 

& 12
& 0.7 & 0.4 & 1.7 & 0.6 & 0.0 & 18.6 & 1.6 & 3.4 \\

& & 16
& 1.3 & 0.4 & 2.6 & 1.8 & 1.6 & 13.6 & 10.4 & 4.5 \\

& & 24
& 10.8 & 44.7 & 32.8 & 42.3 & 44.8 & 21.3 & 44.8 & 34.5 \\

& & 32
& 34.7 & 75.9 & 64.9 & 81.3 & 68.4 & 23.8 & 75.6 & 60.6 \\

& & 48
& 47.6 & 86.7 & 75.8 & 92.7 & 81.7 & 23.9 & 81.3 & 70.0 \\

\bottomrule
\end{tabular}%
}
\vspace{0.1in}
\end{table*}

\clearpage
\newpage
\section{Expert Activation Probabilities in Base DeepSeek-V2-Lite}
\label{appendix:activation_analysis}

\begin{figure}[h]
    \centering
    \includegraphics[width=145mm, height=70mm]{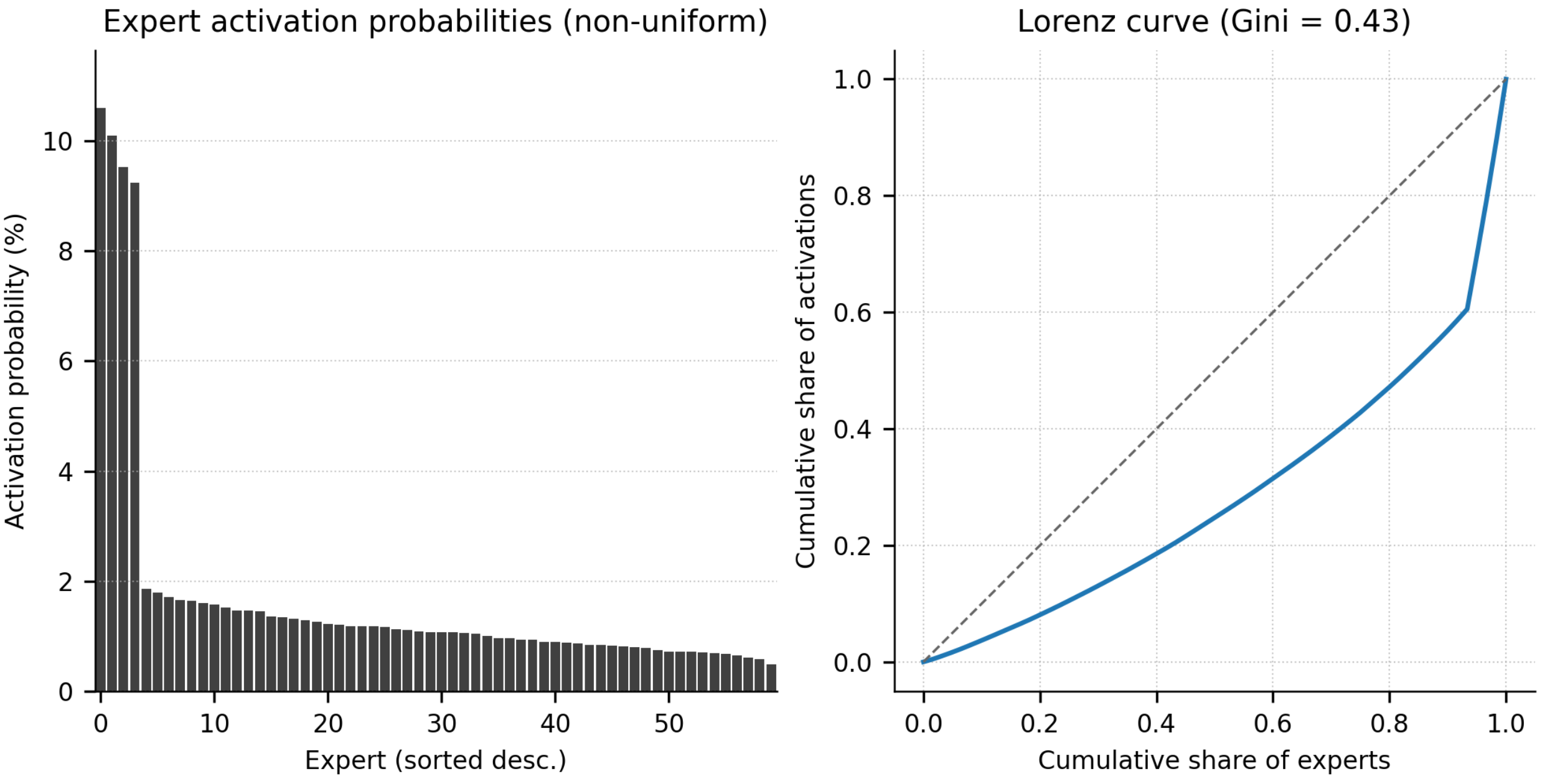}
    \centering
    \caption{Expert activations rate in whole~\textit{math7K} dataset. Expert Activations are tested using the Deepseek-V2-Lite base model.}
    \label{fig:deepseek_base_expert_activate}
\end{figure}

Figure~\ref{fig:deepseek_base_expert_activate} presents the activation statistics of all experts in the DeepSeek-V2-Lite base model when evaluated on the full \textit{math7K} dataset. The left panel illustrates the sorted activation probabilities and reveals a highly non-uniform routing pattern, where a small subset of experts is used disproportionately often, with several exceeding a 10\% activation rate. Beyond this small group, activation frequencies decrease sharply, forming a long tail of under-utilized experts with probabilities falling below 2\% and eventually approaching zero. This distribution indicates that, although the MoE architecture allocates equal computational capacity to every expert, the router naturally converges toward a skewed, winner-take-most configuration.

The right panel shows the corresponding Lorenz curve, which characterizes the cumulative distribution of activations across experts. The curve deviates substantially from the diagonal line representing a perfectly uniform distribution, yielding a Gini coefficient of 0.43. This quantitatively confirms a notable degree of inequality in expert utilization, where a relatively small fraction of experts accounts for the majority of routing decisions.

Despite this concentration of activation mass, pruning experiments demonstrate that activation frequency alone is not a reliable indicator of expert importance. Removing the least-active experts leads to clear accuracy degradation, and even when retaining the top 75\% of experts ranked by activation probability, the model still suffers a performance drop exceeding 10\% on downstream math reasoning benchmarks. Details can be found in Appendix~\ref{appendix:scale_law_tables}. This observation suggests that low-activation experts encode specialized competencies that remain essential for certain categories of problems. Consequently, reducing experts solely based on activation sparsity risks eliminating critical functional diversity, thereby motivating the need for structured or distillation-based compression approaches---such as the ExpertCondenser method proposed in this work---to safely reduce mixture-of-experts models.

\newpage
\section{Ablation Studies}
\label{appendix:ablation-study}

\subsection{Effect of the Number of Condenser Experts}
\label{appendix:r_ablation}

% In this section, we study the impact of the number of condenser experts $r$ on downstream performance. 
% As discussed in the main text, $r$ controls a tradeoff between aggregation capacity and gradient concentration: 
% larger $r$ increases representational capacity but dilutes gradient updates across more experts, 
% while smaller $r$ concentrates gradients but may limit the ability to capture diverse patterns.

% To evaluate this tradeoff, we conduct an ablation study on \textbf{DeepSeek-V2-Lite (16B)} by varying $r \in \{1, 2, 4\}$.
% All other training settings are kept fixed. The results are reported in Table~\ref{tab:r_ablation}.

% We observe that increasing $r$ from $1$ to $2$ improves performance across most benchmarks, 
% suggesting that a single condenser expert is insufficient to capture the diversity of routed expert knowledge. 
% However, further increasing $r$ to $4$, \hector{TODO}.
% This supports our hypothesis that overly large $r$ reduces gradient concentration and weakens the aggregation effect.

% Overall, $r=2$ achieves the best balance between capacity and consolidation, validating the tradeoff described in the main text.

In this section, we study the impact of the number of condenser experts $r$ on downstream performance. 
As discussed in the main text, $r$ controls a tradeoff between aggregation capacity and gradient concentration: 
larger $r$ increases representational capacity but dilutes gradient updates across more experts, 
while smaller $r$ concentrates gradients but may limit the ability to capture diverse patterns.

To evaluate this tradeoff, we conduct an ablation study on \textbf{DeepSeek-V2-Lite (16B)} by varying $r \in \{1, 2, 4\}$.
All other training settings are kept fixed. The results are reported in Table~\ref{tab:r_ablation}.

We observe that increasing $r$ from $1$ to $2$ improves performance across benchmarks, 
suggesting that a single condenser expert is insufficient to capture the diversity of routed expert knowledge. 
However, further increasing $r$ to $4$ yields mixed results: while performance improves on certain datasets (e.g., MultiArith and mawps), it degrades on others (notably AQuA), and does not improve the overall average. 
This behavior is consistent with the tradeoff described above: a larger $r$ increases capacity but reduces gradient concentration per expert, weakening the effectiveness of knowledge aggregation.

Overall, $r=2$ achieves the best balance between capacity and consolidation, validating the tradeoff described in the main text.

\begin{table*}[h!]
\centering
\caption{Ablation study on the number of condenser experts $r$ using DeepSeek-V2-Lite (16B). Results are zero-shot accuracy (\%) on math reasoning benchmarks.}
\label{tab:r_ablation}
\vspace{-0.1in}
\resizebox{\textwidth}{!}{%
  \setlength{\tabcolsep}{1.5mm}%
  \begin{tabular}{p{1.5cm} p{2.8cm} c c c c c c c c c}
    \toprule
    \multirow{2}{*}{\textbf{Dataset}} 
      & \multirow{2}{*}{\textbf{\centering Model}} 
      & \multirow{2}{*}{$\mathbf{r}$}
      & \multirow{2}{*}{\textbf{GSM8k}}
      & \multirow{2}{*}{\textbf{SingleEq}}
      & \multirow{2}{*}{\textbf{SVAMP}} 
      & \multirow{2}{*}{\textbf{MultiArith}} 
      & \multirow{2}{*}{\textbf{AddSub}} 
      & \multirow{2}{*}{\textbf{AQuA}} 
      & \multirow{2}{*}{\textbf{mawps}} 
      & \multirow{2}{*}{\textbf{AVG}} \\
    & & & & & & & & & & \\
    
    \midrule
    \multirow{3}{*}{\textbf{math7k}} 
      & \multirow{3}{*}{\textbf{DeepSeek-V2-Lite}} 
      & $r=1$ & 58.9 & 92.1 & 68.8 & 90.3 & 77.6 & 35.9 & 82.5 & 72.2 \\
      & & $\mathbf{r=2}$ & \textbf{59.4} & \textbf{92.5} & \textbf{69.1} & \textbf{91.5} & \textbf{79.5} & \textbf{36.1} & \textbf{83.6} & \textbf{73.1} \\
      & & $r=4$ & 58.6 & 92.1 & 69.7 & 92.7 & 78.7 & 33.5 & 85.7 & 73.0 \\
      
    \bottomrule
  \end{tabular}%
}
\vspace{-0.1in}
\end{table*}

\subsection{How to Choose Condenser Experts}
\label{appendix:condenser_experts_selection}
In this subsection, we conduct an ablation study to investigate how experts should be selected as shared experts during post-training. 
Table~\ref{tab:bias_condenser_selection} reports the results of comparing two selection strategies: 
(i) choosing high-bias experts and (ii) choosing low-bias experts. 
Across both \textbf{DeepSeek-V2-Lite (16B)} and \textbf{Qwen1.5-MoE (14B)}, we observe that selecting \emph{low-bias experts} consistently leads to stronger downstream performance on math reasoning benchmarks. 
For example, in the \textbf{math7k} setting, low-bias experts achieve higher average accuracy (73.1 vs.\ 72.4 for DeepSeek-V2-Lite and 63.4 vs.\ 61.8 for Qwen1.5-MoE). 
These results suggest that low-bias experts encode more generalizable knowledge, making them more effective as shared experts in MoE post-training.

\paragraph{Proposition (Low-Bias Experts as Effective Aggregators).}

Let $\tilde{s}_i(x) = s_i(x) + b_i$ denote the biased routing logits under auxiliary-free routing, and let $p_i = \mathbb{P}(i \in S(x))$ denote the empirical activation probability of expert $i$. 
While bias $b_i$ influences routing, activation frequency $p_i$ is an empirical property that emerges from both routing logits and data distribution. Therefore, low-bias experts are not strictly equivalent to low-activation experts.

Under standard sparse routing, experts with lower activation probabilities $p_i$ receive fewer gradient updates due to their infrequent selection. Promoting such experts to condenser experts (i.e., enforcing $i \in S(x)$ for all $x$) removes this sparsity factor and increases their expected gradient exposure.

As a result, low-activation experts can transition from under-trained specialists to persistent aggregation nodes that accumulate gradient signals across diverse inputs. In contrast, high-activation experts already receive frequent updates, so promoting them provides limited additional benefit. Randomly selected experts fall between these two extremes, providing moderate but less targeted aggregation.

To validate this, we compare five selection strategies for choosing condenser experts: (i) low-bias, (ii) high-bias, (iii) low-activation, (iv) high-activation, and (v) random selection.

\begin{table*}[h]
\centering
\caption{Evaluation of post-trained models (Zero-Shot results) on downstream Math Reasoning datasets to conduct ablation study on expert selection strategies, including bias-based, activation-based, and random selection.}
\label{tab:bias_condenser_selection}
\vspace{-0.1in}
\resizebox{\textwidth}{!}{%
  \setlength{\tabcolsep}{1.5mm}%
  \begin{tabular}{p{1.5cm} p{2.5cm} c c c c c c c c c c c}
    \toprule
    \multirow{2}{*}{\textbf{Dataset}} 
      & \multirow{2}{*}{\textbf{\centering Model}} 
      & \multirow{2}{*}{\textbf{Model Size}} 
      & \multirow{2}{*}{\textbf{\#Param (Experts)}} 
      & \multirow{2}{*}{\textbf{Selection Type}}
      & \multirow{2}{*}{\textbf{GSM8k}}
      & \multirow{2}{*}{\textbf{SingleEq}}
      & \multirow{2}{*}{\textbf{SVAMP}} 
      & \multirow{2}{*}{\textbf{MultiArith}} 
      & \multirow{2}{*}{\textbf{AddSub}} 
      & \multirow{2}{*}{\textbf{AQuA}} 
      & \multirow{2}{*}{\textbf{mawps}} 
      & \multirow{2}{*}{\textbf{AVG}} \\
    & & & & & & & & & & & & \\
    
    \midrule
    \multirow{5}{*}{\textbf{math7k}} 
      & \multirow{5}{*}{\textbf{DeepSeek-V2-Lite}}
      & \multirow{5}{*}{\textbf{16B}}
      & \multirow{5}{*}{\textbf{2.4B}}
      & high-bias experts & 60.1 & 90.2 & 70.4 & 90.2 & 74.2 & 37.0 & 84.5 & 72.4 \\
      & & & & low-bias experts & 59.4 & 92.5 & 69.1 & 91.5 & 79.5 & 36.1 & 83.6 & 73.1 \\
      & & & & low-activation experts & 59.2 & 91.8 & 69.7 & 90.9 & 78.6 & 35.8 & 83.9 & 72.8 \\
      & & & & high-activation experts & 59.8 & 90.5 & 70.0 & 90.6 & 75.1 & 36.7 & 84.0 & 72.4 \\
      & & & & random experts & 59.2 &91.0 & 70.6 &91.0 & 79.2 &36.2 &83.2& 72.9 \\

    \cmidrule(l){2-13}
    & \multirow{5}{*}{\textbf{Qwen1.5-MoE}} 
    & \multirow{5}{*}{\textbf{14B}}
    & \multirow{5}{*}{\textbf{2.7B}}
      & high-bias experts & 54.7 & 71.0 & 53.8 & 84.6 & 58.8 & 32.8 & 76.8 & 61.8 \\
      & & & & low-bias experts & 57.2 & 74.6 & 55.7 & 86.0 & 61.8 & 33.1 & 75.6 & 63.4 \\
      & & & & low-activation experts & 56.6 & 73.9 & 55.1 & 86.4 & 62.0 & 32.7 & 75.0 & 63.0 \\
      & & & & high-activation experts & 55.0 & 73.8 & 54.2 & 84.9 & 60.4 & 32.5 & 76.2 & 62.4 \\
      & & & & random experts & 56.1 & 72.8 & 55.0 & 85.2 & 60.3 & 32.9 & 75.4 & 62.5 \\
      
    \bottomrule
  \end{tabular}%
}
\vspace{-0.1in}
\end{table*}

\subsection{Gated versus Ungated Condenser Experts}
\label{appendix:gating_ablation}

The ablation in \S~\ref{subsection:expertcondenser_dissecting} varies whether a persistent path exists, not whether that path is gated, so it cannot separate the always-active capacity from the per-token weighting applied to it. We isolate the two by changing a single term of Eq.~\ref{formula:reformulate_moe}: the condenser contribution $g_j(x_t)\,E_j(x_t)$ becomes $c_j\,E_j(x_t)$ with $c_j{=}1$, which is exactly an ungated shared expert. Table~\ref{tab:gating_ablation} reports the control on Qwen1.5-MoE-A2.7B / \texttt{math7k} with $r{=}2$, one epoch, and identical initialization and hyperparameters across rows.
The ungated variant averages $59.3$, matching the $59.4$ of \emph{aux-free+bias}, which has no persistent path at all; gating the same path raises the average to $63.4$. Always-active capacity therefore yields no gain on its own---the improvement comes from keeping the persistent path token-dependent, as argued in \S~\ref{subsection:method_condenser_experts}.

\begin{table*}[h]
\centering

\caption{Gating control on Qwen1.5-MoE-A2.7B / \texttt{math7k} ($r{=}2$, 1 epoch). The last two rows differ only in the condenser weighting: a fixed $c_j{=}1$ versus the router score $g_j(x_t)$. The \emph{aux-free+bias} and gated rows are the corresponding rows of Table~\ref{tab:mix_dataset_results}.}
\label{tab:gating_ablation}
\vspace{-0.05in}
\adjustbox{max width=\textwidth}{%
\setlength{\tabcolsep}{1.5mm}%
\begin{tabular}{l c c c c c c c c}
\toprule
\textbf{Variant} & \textbf{GSM8k} & \textbf{SingleEq} & \textbf{SVAMP} & \textbf{MultiArith} & \textbf{AddSub} & \textbf{AQuA} & \textbf{mawps} & \textbf{AVG} \\
\midrule
SFT & 45.8 & 70.2 & 53.6 & 76.2 & 53.3 & 27.6 & 67.8 & 56.4 \\
aux-free+bias & 48.2 & 76.6 & 52.7 & 80.3 & 59.2 & 26.8 & 71.8 & 59.4 \\
\quad + ungated condenser ($c_j{=}1$) & 52.7 & 71.4 & 50.4 & 82.3 & 60.3 & 26.8 & 71.5 & 59.3 \\
\quad \textbf{+ gated condenser ($g_j(x_t)$)} & \textbf{57.2} & 74.6 & \textbf{55.7} & \textbf{86.0} & \textbf{61.8} & \textbf{33.1} & \textbf{75.6} & \textbf{63.4} \\
\bottomrule
\end{tabular}%
}
\end{table*}

\clearpage
\newpage

\section{Training Stability Analysis of ExpertCondenser}
\label{appendix:training_stability}

\begin{figure}[h]
    \centering
    \includegraphics[width=120mm]{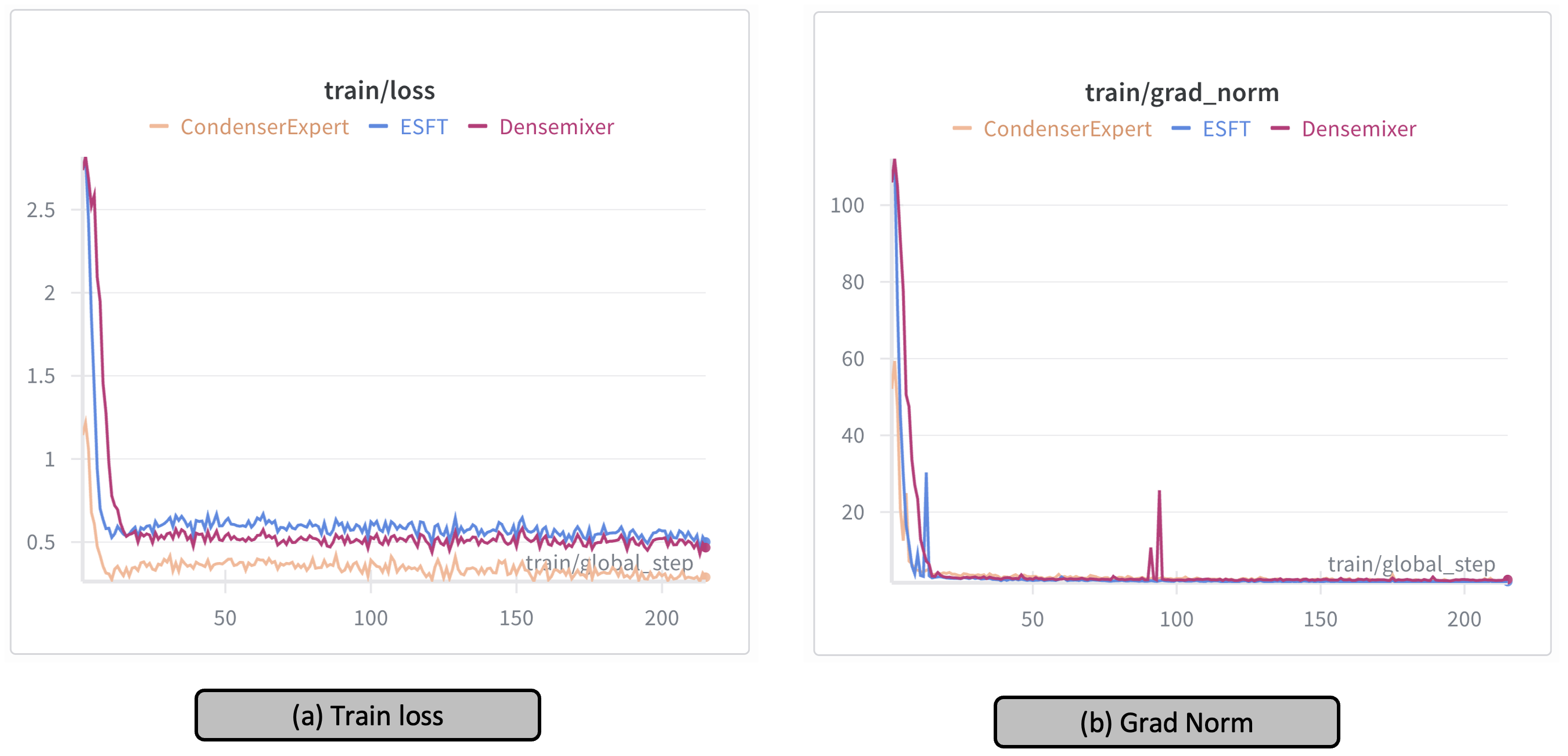}
    \caption{Training stability comparison among \textsc{ExpertCondenser}, ESFT, and DenseMixer. 
    (a) Training loss curves, where \textsc{ExpertCondenser} converges more quickly and reaches a lower final loss. 
    (b) Gradient-norm evolution, showing that \textsc{ExpertCondenser} maintains smooth, stable gradients without the spikes observed in DenseMixer and ESFT.}
    \label{fig:training_stability}
\end{figure}

Figure~\ref{fig:training_stability} presents the training dynamics of \textsc{ExpertCondenser} compared with the ESFT and DenseMixer baselines when post-training the \textsc{GPT-OSS} model on the \textit{math7K} dataset. The left panel plots the training-loss curves, while the right panel reports the corresponding gradient norms over the full optimization trajectory.

Across training, \textsc{ExpertCondenser} displays markedly improved stability and convergence efficiency relative to the baselines. Its loss curve decreases rapidly during early optimization and consistently settles at a lower final value than both ESFT and DenseMixer. This indicates more effective optimization dynamics and better alignment with the target reasoning distribution. ESFT converges to a higher asymptotic loss, whereas DenseMixer converges slower.

The gradient-norm behavior further highlights this stability advantage. After the initial warm-up phase, \textsc{ExpertCondenser} maintains smooth and well-bounded gradients, free of the high-magnitude spikes observed in DenseMixer and the early instability exhibited by ESFT. The absence of gradient bursts suggests that \textsc{ExpertCondenser} operates over a more stable gradient landscape, which contributes to improved convergence and reduced optimization volatility.

Taken together, these results demonstrate that \textsc{ExpertCondenser} not only delivers better downstream accuracy but also enhances the stability and reliability of the training process compared to existing post-training approaches.

\clearpage
\newpage

\section{Formalizing the Motivation Behind Condenser Experts}
\label{appendix:condenser_theory}

Mixture-of-Experts (MoE) architectures rely on two forms of expert capacity:
(i) \emph{sparse routed experts}, selected dynamically via top-$k$ routing, and 
(ii) \emph{static shared experts}, which are always active and contribute a fixed residual transformation.

The static shared experts contribute
\[
\sum_{k=1}^{\hat{n}} E_k(\mathbf{x}_t),
\]
independent of routing decisions. While this provides persistent capacity, these experts are ungated and therefore cannot adapt their contribution to the input or interact with routing dynamics.

To address this limitation, we introduce \textbf{condenser experts}, which are \emph{always selected} yet remain \emph{gated per token}. 
Let $J \subset \{1, \dots, n\}$ denote the set of condenser experts with $|J| = r$. 
For each token $\mathbf{x}_t$, the active expert set is defined as
\[
S(\mathbf{x}_t) = J \cup \operatorname{TopK}_{i \notin J} \left( \tilde{s}_i(\mathbf{x}_t),\, k - r \right).
\]

Following Eq.~\ref{formula:reformulate_moe}, the MoE output is
\[
h'_t =  
\underbrace{\sum_{j \in J} g_j(\mathbf{x}_t)\, E_j(\mathbf{x}_t)}_{\text{gated condenser experts}}
+
\underbrace{\sum_{i \in S(\mathbf{x}_t)\setminus J} g_i(\mathbf{x}_t)\, E_i(\mathbf{x}_t)}_{\text{selected sparse experts}}
+
\underbrace{\sum_{k=1}^{\hat{n}} E_k(\mathbf{x}_t)}_{\text{static shared experts}}.
\]

This formulation highlights that condenser experts define a \emph{persistent but input-dependent computation path}, in contrast to static shared experts which are input-independent.

\subsection{Router Stability and Anti-Collapse Behavior}

Condenser experts also improve routing stability. The gradient with respect to router parameters $\phi$ is
\[
\nabla_\phi \mathcal{L}(\mathbf{x})
=
\sum_{i \in S(\mathbf{x})}
\nabla_\phi g_i(\mathbf{x})\, E_i(\mathbf{x}),
\]
where $J \subset S(\mathbf{x})$ guarantees that part of this gradient is always present.

This acts as a structural regularizer by:
(i) reducing routing variance,
(ii) ensuring consistent gradient flow, and
(iii) mitigating collapse of long-tailed experts.

Unlike static shared experts, which bypass routing, condenser experts propagate both forward and backward signals through the router.

\subsection{Capacity Constraints and Variance Reduction}

By enforcing $J \subset S(\mathbf{x}_t)$, the MoE layer incorporates a structured capacity constraint:
\[
|S(\mathbf{x}_t)| = k, \quad J \subset S(\mathbf{x}_t).
\]

Under stochastic routing, the output variance can be decomposed as
\[
\mathrm{Var}[h'_t] =
\mathrm{Var}\!\left[
\sum_{i \in S(\mathbf{x}_t)\setminus J} g_i(\mathbf{x}_t)\,E_i(\mathbf{x}_t)
\right]
+
\mathrm{Var}\!\left[
\sum_{j \in J} g_j(\mathbf{x}_t)\,E_j(\mathbf{x}_t)
\right].
\]

The second term is invariant to routing selection randomness, since condenser experts are always active. 
This reduces stochasticity induced by dynamic routing and leads to more stable optimization.

\clearpage
\newpage

\section{Hyper-parameters}

\subsection{Math7K Dataset and Math14K Dataset}

In this subsection, we perform an ablation study to verify whether the number of fine-tuning epochs is sufficient for convergence. 
Since training efficiency and stability are critical in post-training large MoE models, it is important to ensure that extending training does not yield further improvements or lead to overfitting. 
We therefore evaluate the performance of \textbf{DeepSeek-V2-Lite (16B)} and \textbf{OLMoE (7B)} under the \textbf{math7k} and \texttt{math14k} benchmark with \textbf{ESFT} fine-tuning for 1, 2, and 3 epochs. 
The results are reported in Table~\ref{tab:epoch_num_deepseek} and~\ref{tab:epoch_num_olmoe}.

Overall, these findings confirm that our main experiments are conducted with models that have already converged, and that increasing the number of training epochs does not lead to meaningful gains.

\begin{table*}[htbp]
\centering
\caption{Evaluation of SFT model Zero-Shot Results on downstream math reasoning tasks after fine-tuning with \textit{Math-14K}, including SingleEQ, MultiArith, AddSub, GSM8K, SVAMP, and AQuA.}
\label{tab:epoch_num_olmoe}
\vspace{-0.1in}
\resizebox{\textwidth}{!}{%
  \setlength{\tabcolsep}{1.5mm}%
  \begin{tabular}{p{3cm} c c c c c c c c c c c}
    \toprule
    \multirow{2}{*}{\textbf{Model}} 
      & \multirow{2}{*}{\textbf{Model Size}} 
      & \multirow{2}{*}{\textbf{\#Param (Experts)}} 
      & \multirow{2}{*}{\textbf{Distill Type}}
      & \multirow{2}{*}{\textbf{GSM8k}}
      & \multirow{2}{*}{\textbf{SingleEq}}
      & \multirow{2}{*}{\textbf{SVAMP}} 
      & \multirow{2}{*}{\textbf{MultiArith}} 
      & \multirow{2}{*}{\textbf{AddSub}} 
      & \multirow{2}{*}{\textbf{AQuA}} 
      & \multirow{2}{*}{\textbf{mawps}} 
      & \multirow{2}{*}{\textbf{AVG}} \\
    & & & & & & & & & & \\
    \midrule
    
    %===== ESFT Models =====
    \multirow{3}{*}{\textbf{OLMoE}}
    & \multirow{3}{*}{\textbf{7B}}
    & \multirow{3}{*}{\textbf{1B}}
   & \centering SFT-1epoch & 55.7 & 76.2 & 58.8 & 71.2 & 62.8 & 28.3 & 64.3 & 59.6 \\
  & & & \centering SFT-2epoch & 52.8 & 78.3 & 59.1 & 71.8 & 63.3 & 29.1 & 68.9 & 60.5 \\
    & & & \centering SFT-3epoch &  52.6 & 77.2 & 57.8 & 72.7 & 64.6 & 31.9 & 70.1 & 60.9 \\
    % \textbf{OLMOE}            & \textbf{7B}  & \centering ESFT & \textbf{1B}   & 64.4 & 77.0 & 68.9 & 81.8 & 64.1 & 30.7 & 74.8 & 65.9 \\
    \bottomrule
  \end{tabular}%
}
\vspace{0.5in}
\end{table*}

\begin{table*}[htbp]
\centering
\caption{Evaluation of SFT model Zero-Shot Results on downstream math reasoning tasks after fine-tuning with \textit{Math-7K}, including SingleEQ, MultiArith, AddSub, GSM8K, SVAMP, and AQuA.}
\label{tab:epoch_num_deepseek}
\vspace{-0.1in}
\resizebox{\textwidth}{!}{%
  \setlength{\tabcolsep}{1.5mm}%
  \begin{tabular}{p{3cm} c c c c c c c c c c c}
    \toprule
    \multirow{2}{*}{\textbf{Model}} 
      & \multirow{2}{*}{\textbf{Model Size}} 
      & \multirow{2}{*}{\textbf{\#Param (Experts)}} 
      & \multirow{2}{*}{\textbf{Distill Type}}
      & \multirow{2}{*}{\textbf{GSM8k}}
      & \multirow{2}{*}{\textbf{SingleEq}}
      & \multirow{2}{*}{\textbf{SVAMP}} 
      & \multirow{2}{*}{\textbf{MultiArith}} 
      & \multirow{2}{*}{\textbf{AddSub}} 
      & \multirow{2}{*}{\textbf{AQuA}} 
      & \multirow{2}{*}{\textbf{mawps}} 
      & \multirow{2}{*}{\textbf{AVG}} \\
    & & & & & & & & & & \\
    \midrule
    
    %===== ESFT Models =====
    \multirow{3}{*}{\textbf{DeepSeek-V2-Lite}}
    &\multirow{3}{*}{\textbf{16B}}
    &\multirow{3}{*}{\textbf{2.4B}}
& \centering ESFT-1epoch & 54.1 & 88.0 & 65.3 & 83.7 & 72.7 & 26.8 & 79.4 & 67.1 \\
      & & &\centering ESFT-2epoch & 58.6 & 80.9 & 65.8 & 90.7 & 62.3 & 27.6& 76.1 & 66.0 \\
      & & & \centering ESFT-3epoch & 58.2 & 75.8 & 65.2 & 89.0 & 56.5 & 29.5 & 73.5 & 64.0 \\
    \midrule
    \multirow{3}{*}{\textbf{OLMoE}}
    & \multirow{3}{*}{\textbf{7B}}
    & \multirow{3}{*}{\textbf{1B}}
    & \centering SFT-1epoch  & 57.0 & 78.5 & 58.6 & 72.0 & 64.3 & 28.3 & 76.1 & 62.1 \\
    & & & \centering SFT-2epoch  & 53.8 & 70.9 & 55.7 & 65.0 & 61.5 & 31.5 & 69.7 & 58.3 \\
    & & & \centering SFT-3epoch  & 50.3 & 69.3 & 49.5 & 54.5 & 59.2 & 27.6 & 59.2 & 52.8 \\
    % \textbf{OLMOE}            & \textbf{7B}  & \centering ESFT & \textbf{1B}   & 62.2 & 75.2 & 68.0 & 93.3 & 58.2 & 28.7 & 73.1 & 65.5 \\
    \bottomrule
  \end{tabular}%
}

\vspace{0.5in}
\end{table*}

\clearpage
\newpage
\subsection{Gamma for Bias Update}

\begin{table*}[htbp]
\centering
\caption{Evaluation of different post-trained Qwen1.5 model Zero-Shot Results on downstream math reasoning tasks with different gamma settings after fine-tuning with \textit{Math-7K}, including SingleEQ, MultiArith, AddSub, GSM8K, SVAMP, and AQuA.}
\label{tab:qwen_gamma_res}
\vspace{-0.1in}
\resizebox{\textwidth}{!}{%
  \setlength{\tabcolsep}{1.5mm}%
  \begin{tabular}{p{3cm} c c c c c c c c c c c}
    \toprule
    \multirow{2}{*}{\textbf{Model Gamma}} 
      & \multirow{2}{*}{\textbf{Model Size}} 
      & \multirow{2}{*}{\textbf{Distill Type}}
      & \multirow{2}{*}{\textbf{\#Param (Experts)}} 
      & \multirow{2}{*}{\textbf{GSM8k}}
      & \multirow{2}{*}{\textbf{SingleEq}}
      & \multirow{2}{*}{\textbf{SVAMP}} 
      & \multirow{2}{*}{\textbf{MultiArith}} 
      & \multirow{2}{*}{\textbf{AddSub}} 
      & \multirow{2}{*}{\textbf{AQuA}} 
      & \multirow{2}{*}{\textbf{mawps}} 
      & \multirow{2}{*}{\textbf{AVG}} \\
    & & & & & & & & & & \\
    \midrule
    
    %===== Qwen1.5 Models Gamma =====
    % \textbf{SFT-Gamma-1e-4} & \textbf{14B} & \centering aux-free-loss & \textbf{2.7B} & - & - & - & - & - & - & - & - \\

    % \midrule
    \textbf{G-1e-6} & \textbf{14B} & \centering aux-free-loss & \textbf{2.7B} & 48.6 & 76.8 & 50.0 & 81.8 & 59.6 & 29.6 & 70.6 & 59.6 \\
    \textbf{G-1e-5} & \textbf{14B} & \centering aux-free+bias & \textbf{2.7B} & 48.8 & 76.4 & 50.2 & 81.6 & 59.8 & 29.7 & 70.8 & 59.6 \\
    \textbf{G-5e-5} & \textbf{14B} & \centering aux-free+bias & \textbf{2.7B} & 47.8 & 76.6 & 50.2 & 81.8 & 60.3 & 29.9 & 70.6 & 59.6 \\
    \textbf{G-1e-4} & \textbf{14B} & \centering aux-free+bias & \textbf{2.7B} & 48.2 & 76.6 & 52.7 & 80.3 & 59.2 & 26.8 & 71.8 & 59.4 \\
    \textbf{G-1e-3} & \textbf{14B} & \centering aux-free+bias & \textbf{2.7B} & 47.7 & 74.2 & 50.6 & 83.2 & 59.0 & 30.7 & 67.2 & 58.9 \\
    \textbf{G-3e-3} & \textbf{14B} & \centering aux-free+bias & \textbf{2.7B} & 31.8 & 58.5 & 37.9 & 65.2 & 39.5 & 28.3 & 56.7 & 45.4 \\
    \textbf{G-5e-3} & \textbf{14B} & \centering aux-free+bias & \textbf{2.7B} & 15.3 & 32.7 & 26.0 & 47.5 & 29.4 & 21.7 & 33.2 & 29.4 \\
    \textbf{G-1e-2} & \textbf{14B} & \centering aux-free+bias & \textbf{2.7B} & 7.2 & 17.3 & 14.7 & 28.2 & 15.7 & 17.7 & 18.5 & 17.0 \\

    \bottomrule
  \end{tabular}%
}
\vspace{-0.1in}
\end{table*}

\begin{table*}[htbp]
\centering
\caption{Evaluation of different post-trained Deepseek-v2-lite model Zero-Shot Results on downstream math reasoning tasks with different gamma settings after fine-tuning with \textit{Math-7K}, including SingleEQ, MultiArith, AddSub, GSM8K, SVAMP, and AQuA.}
\label{tab:deepseek_gamma_res}
\vspace{-0.1in}
\resizebox{\textwidth}{!}{%
  \setlength{\tabcolsep}{1.5mm}%
  \begin{tabular}{p{3cm} c c c c c c c c c c c}
    \toprule
    \multirow{2}{*}{Model Gamma}
      & \multirow{2}{*}{Model Size}
      & \multirow{2}{*}{Distill Type}
      & \multirow{2}{*}{\#Param (Experts)}
      & \multirow{2}{*}{GSM8k}
      & \multirow{2}{*}{SingleEq}
      & \multirow{2}{*}{SVAMP}
      & \multirow{2}{*}{MultiArith}
      & \multirow{2}{*}{AddSub}
      & \multirow{2}{*}{AQuA}
      & \multirow{2}{*}{mawps}
      & \multirow{2}{*}{AVG} \\
    & & & & & & & & & & \\
    \midrule
    \textbf{G-1e-6} & \textbf{14B} & \centering aux-free-loss & \textbf{2.7B} & 56.2 & 89.4 & 68.9 & 86.2 & 73.2 & 35.2 & 79.0 & 69.8 \\
    \textbf{G-1e-5} & \textbf{14B} & \centering aux-free-loss & \textbf{2.7B} & 56.7 & 89.6 & 69.8 & 87.8 & 74.0 & 35.8 & 79.6 & 70.5 \\
    \textbf{G-5e-5} & \textbf{14B} & \centering aux-free-loss & \textbf{2.7B} & 58.6 & 90.6 & 70.2 & 88.6 & 74.4 & 36.2 & 80.6 & 71.3 \\
    \textbf{G-1e-4} & \textbf{14B} & \centering aux-free-loss & \textbf{2.7B} & 58.8 & 90.7 & 69.3 & 88.7 & 74.2 & 36.1 & 80.3 & 71.2 \\
    \textbf{G-1e-3} & \textbf{14B} & \centering aux-free-loss & \textbf{2.7B} & 43.4 & 78.7 & 62.8 & 80.5 & 60.5 & 25.2 & 71.0 & 60.3 \\
    \bottomrule
  \end{tabular}%
}
\vspace{-0.1in}
\end{table*}

\end{document}